\documentclass[journal]{IEEEtran}
\usepackage{cite}
\usepackage{amsmath,amssymb,amsfonts}
\usepackage{graphicx}
\usepackage{textcomp}
\usepackage{rotating}
\usepackage{soul}

\usepackage{bbm}
\usepackage{gensymb}
\usepackage{algorithmic}
\usepackage[ruled,vlined]{algorithm2e}
\usepackage[normalem]{ulem}
\usepackage{multirow}
\usepackage{xcolor}

\usepackage{romannum}

\def\BibTeX{{\rm B\kern-.05em{\sc i\kern-.025em b}\kern-.08em
    T\kern-.1667em\lower.7ex\hbox{E}\kern-.125emX}}
\begin{document}
\title{CANet: Context Aware Network for Brain Glioma Segmentation}
\author{Zhihua Liu, Lei Tong, Long Chen, Feixiang Zhou, Zheheng Jiang, Qianni Zhang, Yinhai Wang, Caifeng Shan, ~\IEEEmembership{Senior Member,~IEEE,} Ling Li and Huiyu Zhou
\thanks{Zhihua Liu, Lei Tong, Long Chen, Feixiang Zhou, Zheheng Jiang are with the School of Informatics, University of Leicester, Leicester, United Kingdom, LE1 7RH. E-mail: zl208@leicester.ac.uk}
\thanks{Qianni Zhang is with the School of Electronic Engineering and Computer Science, Queen Mary University of London, London, United Kingdom}
\thanks{Yinhai Wang is with Biopharmaceutical R\&D, AstraZeneca, Unit 310 - Darwin Building, Cambridge Science Park, Milton Road, Cambridge, United Kingdom, CB4 0WG}
\thanks{Caifeng Shan is with Philips Research, High Tech Campus, 5656 AE, Eindhoven, The Netherlands}
\thanks{Ling Li is with the School of Computing, University of Kent, United Kingdom}
\thanks{Huiyu Zhou is with the School of Informatics, University of Leicester, Leicester, United Kingdom, LE1 7RH. Corresponding author, E-mail: hz143@leicester.ac.uk}}

\maketitle

\begin{abstract}
Automated segmentation of brain glioma plays an active role in diagnosis decision, progression monitoring and surgery planning. Based on deep neural networks, previous studies have shown promising technologies for brain glioma segmentation. However, these approaches lack powerful strategies to incorporate contextual information of tumor cells and their surrounding, which has been proven as a fundamental cue to deal with local ambiguity. In this work, we propose a novel approach named Context-Aware Network (CANet) for brain glioma segmentation. CANet captures high dimensional and discriminative features with contexts from both the convolutional space and feature interaction graphs. We further propose context guided attentive conditional random fields which can selectively aggregate features. We evaluate our method using publicly accessible brain glioma segmentation datasets BRATS2017, BRATS2018 and BRATS2019. The experimental results show that the proposed algorithm has better or competitive performance against several State-of-The-Art approaches under different segmentation metrics on the training and validation sets.
\end{abstract}


\section{Introduction}
\label{sec:introduction}
\IEEEPARstart{G}{lioma} is one of the most prevalent types of adult brain tumor with fateful health damage impacts and high mortality \cite{black2005cancer}. To provide sufficient evidence for early diagnosis, surgery planning and post-surgery observation, Magnetic Resonance Imaging (MRI) with multi-modalities (e.g. T1, T1 with contrast-enhanced (T1ce), T2 and Fluid Attenuation Inversion Recover (FLAIR)) is a widely used diagnosis technique to provide reproducible and non-invasive measurement, including structural, anatomical and functional characteristics.

\begin{figure}
    \centering
    \includegraphics[width=0.5\textwidth]{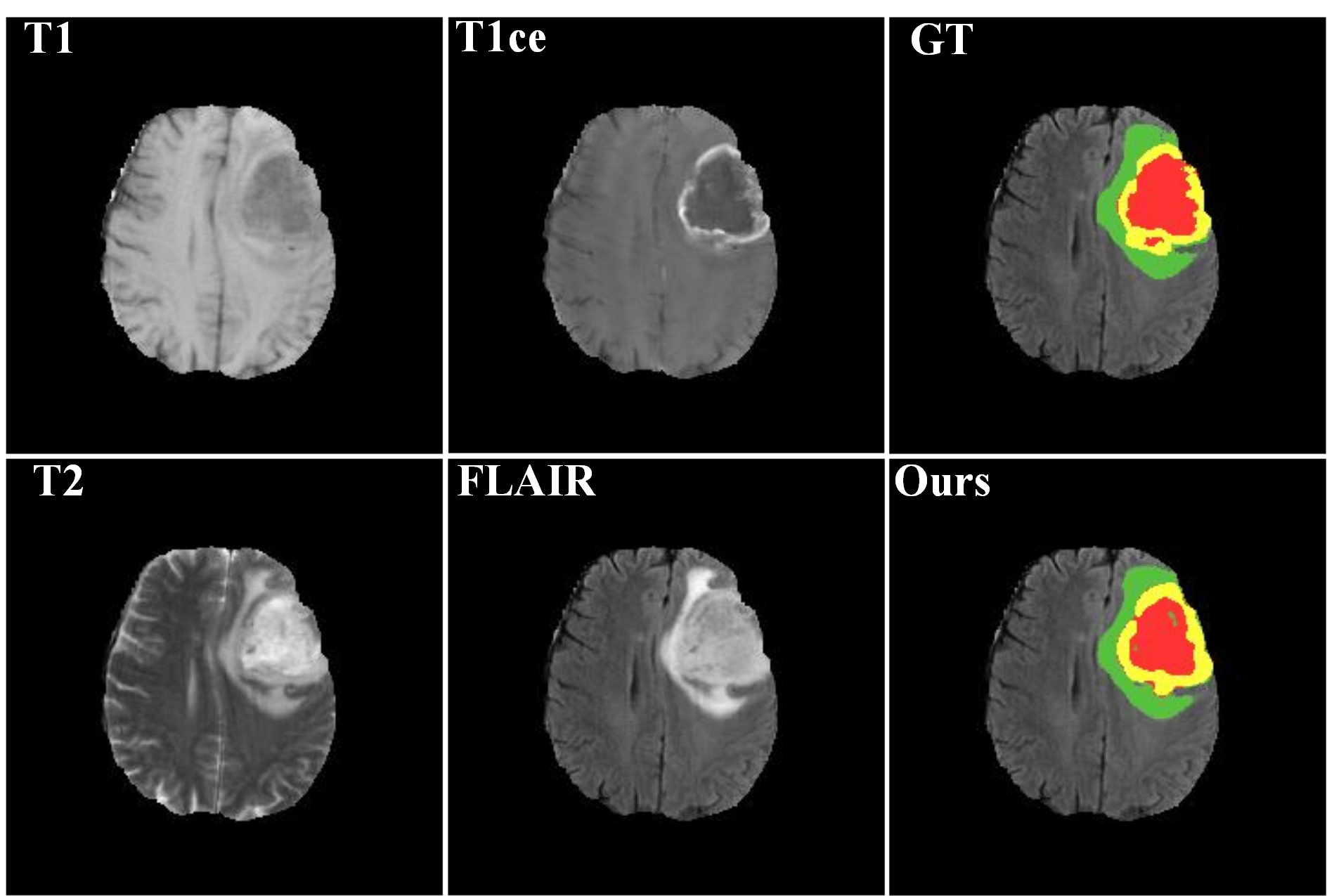}
    \caption{Examples of multi-modality image slices from BraTS17 with the ground-truth and our segmentation results. In this figure, green represents GD-Enhancing Tumor (numerical label 2), yellow represents Pertumoral Edema (numerical label 1) and red represents Necrotic and Non-Enhancing Tumor Core (NCR$\backslash$ECT, numerical label 4).}
\label{fig:task}
\end{figure}

\begin{figure*}[h]
    \centering
    \includegraphics[width=\textwidth]{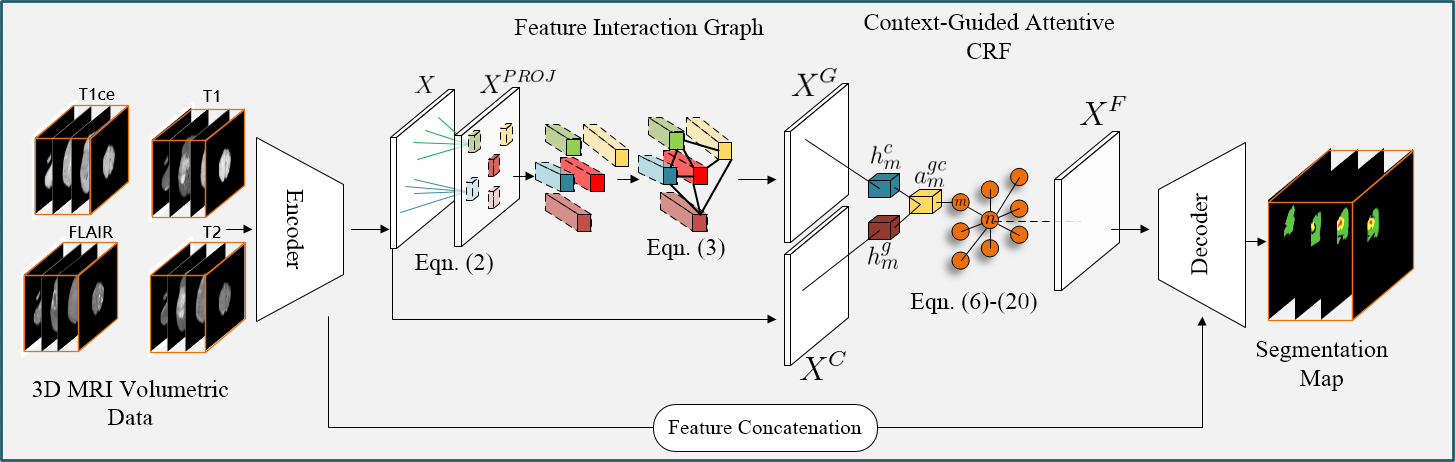}
    \caption{The architecture of the proposed context aware network.}
    \label{fig:totalsystem}
\end{figure*}

Medical image segmentation provides fundamental guidance and quantitative assessment for medical professionals to achieve disease diagnosis, tumor growth monitoring, planning treatment and follow-up services \cite{shen2017deep, mang2020integrated}. Fig. \ref{fig:task} shows an overview of the brain glioma segmentation task. However, manual segmentation requires professional expertise and tends to be time consuming and labour intensive. Previous methods on automated brain glioma segmentation were based on traditional machine learning algorithms \cite{bauer2012segmentation, subbanna2012probabilistic, shin2012hybrid, festa2013automatic}, which strongly rely on hand-crafted features, such as textures \cite{reza2013multi} and local histograms \cite{goetz2014extremely}. However, finding the best hand-crafted features or optimal feature combinations is impracticable. In recent years, deep learning techniques, especially deep convolutional neural network (DCNN), have been deployed to effectively learn high dimensional discriminative features from data and widely used on various medical imaging tasks \cite{litjens2017survey}.

Inter-class ambiguity is a common issue in brain glioma segmentation. This issue makes it hard to achieve accurate dense voxel-wise segmentation if we only consider isolated voxels, as different classes' voxels may share similar intensity values or feature representations. To address this issue, we aim to learn relational information between glioma cells and their surroundings by exploring their feature interaction graphs. We here propose a context-aware network, namely CANet, to achieve accurate dense voxel-wise brain glioma segmentation in MRI images. Our contributions in this work are summarised below:

\begin{itemize}
    \item We propose a novel brain glioma segmentation approach by introducing feature interaction graph reasoning as a parallel auxiliary branch to model the relation between glioma cells and their surroundings. The intermediate feature representations are further exploited and aggregated within a customized context guided attentive conditional random field (CGA-CRF) framework. To our knowledge, this is the first practice on brain glioma segmentation that incorporates relational information from the generated features.
    \item We formulate the mean-field approximation of the inferences in the proposed CGA-CRF as a convolution operation, whereas CGA-CRF is implemented as sequential deep neural networks layers. Our formulation demonstrates the generalization capability of the proposed CGA-CRF that can be embedded within any deep neural architecture seamlessly to achieve end-to-end training.
    \item We conduct extensive evaluations to demonstrate that our proposed approach outperforms several State-of-The-Art methods under different evaluation metrics on the Multimodal Brain Tumor Image Segmentation Challenge (BraTS) datasets, i.e. BraTS2017, BraTS2018 and BraTS2019.
\end{itemize}

\section{Related Work}
\label{Related Work}
We construct our novel brain glioma segmentation approach upon recent successes of deep neural networks and probabilistic graphical models. Below, we briefly review related methods categorising them into three sub-areas, \textit{i.e.} brain glioma segmentation, semantic segmentation using conditional random field combined with convolutional neural network, and graph neural network in medical image analysis.

\textbf{Brain Glioma Segmentation.} \textcolor{red}{}Early research works on brain glioma segmentation mainly used traditional machine learning algorithms, such as clustering \cite{shin2012hybrid}, random decision forests \cite{festa2013automatic}, Bayesian models \cite{corso2008efficient} and graph-cuts \cite{wels2008discriminative}. Shin \cite{shin2012hybrid} used sparse coding for generating edema features and K-means for clustering tumor voxels. However, how to optimise the size of coding dictionary is an intractable problem. Pereira et al. \cite{pereira2016brain} classified voxels' labels using random decision forests, which heavily relied on hand-crafted features and complicated postprocessing. Corso et al. \cite{corso2008efficient} used a Bayesian formulation for incorporating soft model assignments into the affinity calculation. This method considered the weighted aggregation of multi-scale features, but ignored the relationship between different scales. Wels et al. \cite{wels2008discriminative} proposed a graph-cut based method to learn optimal graph representations for tumor segmentation, resulting in a superior performance. However, this method required a prolonged inference duration for dense segmentation tasks, as the number of vertices in its graph is proportional to the number of the voxels. Konukoglu et al. \cite{konukoglu2010extrapolating} and Menze et al. \cite{menze2011generative} incorporated a reaction-diffusion based biophysical tumor growth framework for glioma segmentation. The former focused on constructing the irradiation invasion margin at a single time instance while the latter focused on formalizing the macroscopic tumor growth model on longitude data. However, both methods required detailed domain knowledge to define the parameters, which limits generalization performance of their methods.

Promising achievements have been made on multi-modal MRI brain glioma segmentation using DCNN. Zikic et al. \cite{zikic2014segmentation} was one of the earliest works that apply DCNN onto brain glioma segmentation. Havaei et al \cite{havaei2017brain} proposed an improved DCNN by using muconvolutional kernels to extract local and global features. Zhao et al. \cite{zhao2018deep} proposed a modified fully convolutional network (FCN) \cite{long2015fully} with conditional random fields as post-processing module for refining brain glioma segmentation. Dong et al. \cite{dong2017automatic} proposed a modified U-Net \cite{ronneberger2015u} for brain glioma segmentation. These previous works used 2D convolutional kernels on 2D image slices generated from the original 3D volumetric data. Methods of using 2D slices decrease the number of the used parameters and require less memory. However, this procedure also leads to the missing of spatial contexts. To minimise information loss and exploit the evidence of adjacent slices, Lyksborg et al. \cite{lyksborg2015ensemble} ensembled three 2D CNNs on three orthogonal 2D patches.

To fully make use of 3D information, recent works applied 3D convolutional kernels onto volumetric data. Kamnitsas et al. \cite{kamnitsas2017efficient} proposed a two pathway 3D DCNN, followed by a dense CRF, for brain glioma segmentation. Authors of \cite{kamnitsas2017efficient} further extended the work using model ensembling \cite{kamnitsas2017ensembles}. Their proposed system EMMA ensembled the models from fully convolutional network (FCN) \cite{long2015fully}, U-Net \cite{ronneberger2015u} and DeepMedic \cite{kamnitsas2016deepmedic}. To avoid over-fitting problems in 3D voxel-level segmentation on limited training datasets, Myronenko \cite{myronenko20183d} proposed a 3D DCNN with an additional variational autoencoder to regularise the decoder by reconstructing the input image.

Medical image datasets (e.g. BraTS) usually have imbalance and inter-class interference problems. To address this issue whilst maintaining segmentation performance, Chen et al. \cite{chen2018focus} and Wang et al. \cite{wang2017automatic} both applied cascaded network structures for segmenting brain glioma, where the input of the inner region segmentation network is the output of the outer region segmentation network. However, cascaded cropping networks mainly focus on one tumor region in one particular network stage and cannot infer the relationship between different tumor regions.

\textbf{Semantic segmentation using conditional random field and convolutional neural network.} Brain glioma segmentation, along with the generic semantic segmentation, aims to assign each pixel with a specific semantic label. Among all the traditional machine learning methods, probabilistic graphical model, especially conditional random field has been considered as one of the most successful representation methods for solving semantic segmentation tasks \cite{kumar5hierarchical, toyoda2008random, kohli2009robust, arnab2016higher}. In order to learn complex unary and pairwise potentials in the end-to-end fashion, recently proposed works focused on solving CRF using deep neural networks. Zheng et al. \cite{zheng2015conditional} formulated the mean-field updates of CRF as a recurrent neural network (RNN). Thus CRF parameters can be updated iteratively with back-propagation. There are mainly two drawbacks in the aforementioned works. First, most of them exploit CRF as a post-processing component to refine the segmentation labels, and hence cannot regulate the feature learning procedure. Second, most of these works allow CRF to smooth the segmentation map by encouraging spatial coherence \cite{chen2018end}. Different from these works, our proposed CGA-CRF mainly contributes to feature aggregation, jointly trained with the network backbone. Moreover, our proposed CGA-CRF considers the contextual information extracted from both the convolution/interaction spaces and the contextual attributes can be generated from a learned attention mechanism.

\textbf{Graph Neural Network in Medical Image Analysis.} In recent years, graph neural network  attracts the attention of researchers in medical image analysis for relational information learning \cite{shanahan2020explicitly}. Parisot et al. \cite{parisot2018disease} constructed a population graph for degenerative disease classification where each node represents features from an individual patient. Li et al. \cite{li2020structured} proposed a topology-adaptive graph neural network for landmark detection with applications on X-ray images. Chao et al. \cite{chao2020lymph} introduced a graph neural network to model the relationship between inter lymph nodes for gross tumor detection. Other research works also applied graph neural network directly onto structured data. For example, Chen et al. \cite{chen2020automated} applied graph neural network for intracranial artery labeling using cerebral artery map data. Li et al. \cite{li2020pooling} fed the brain functional graph data into graph neural network for brain biomarker analysis. The aforementioned research works treated pre-localised regions of the image as graph vertices, which greatly limits the generalisation of these methods over different datasets. Our feature interaction graph treats the feature instances as vertices, which provides the relational learning ability and data independent transferability.

\section{Proposed Method}
\label{Proposed Method}
In this section, we describe our proposed CANet for voxel-wise brain glioma segmentation. We first describe the proposed feature interaction graph in detail. Then we introduce the novel feature fusion module, CGA-CRF, which selectively aggregates the features generated from different contexts and learns to render optimal features. Finally, the formulation of the mean-field updates in CGA-CRF as sequential convolutional operations is described, enabling the network to achieve end-to-end training. The proposed segmentation framework is illustrated in Fig. \ref{fig:totalsystem}. Supplementary B summarises the training steps of our proposed CANet.

Different from previous works, our proposed CANet can implicitly capture long-range relational information by reasoning on the feature interaction graph, which have not been fully studied in the literature. Both contexts (feature interaction graph and convolution) utilise the intermediate feature map $X\in \mathbb{R}^{N \times C}$ derived from the shared encoder backbone as input, where $N = H\times W \times D$ is the total number of the feature instances in the intermediate feature map. $C$ is the number of the feature dimension. The graph context generates representations in the feature interaction graph space $X^{G} \in \mathbb{R}^{N \times C}$ and the convolution generates a coordinate space representation $X^{C}\in \mathbb{R}^{N \times C}$.

The main concept behind the design of CGA-CRF is to generate an optimal segmentation map $H \in \mathcal{H}$ associated with an MRI image $I \in \mathcal{I}$ by exploiting the relationship between the final representation $X^{F} \in \mathbb{R}^{N \times C}$ and the intermediate feature representation $X$ with auxiliary long-range relational information $X^{G}$, generated from the interaction space with its convolution features $X^{C}$. Different from direct concatenation $X^{F} = \textit{concat}(X, X^{G}, X^{C})$ or element-wise summation $X^{F} = X + X^{G} + X^{C}$, we aim to learn a set of latent feature representations $X^{F}$ through a novel conditional random field. Since $X^{C}$ and $X^{G}$ may contribute differently during the learning $X^{F}$, we adopt the idea of an attention mechanism and generalise it to a gate node of CRFs. The gate node can regulate the information flow and discover the relevance between different contexts and latent features.
\subsection{Context Guided Feature Extraction}
\label{HCG-FA}
\subsubsection{Graph Context}
\textbf{Projection with Adaptive Sampling} We first use the collected feature map to create a feature interaction space by constructing an interaction graph $ G = \{V,  E, A\}$, where $ V$ represents the set of nodes in the interaction graph, $ E$ represents the edges between the interaction nodes and $A$ represents the adjacency matrix. Given a learned high dimensional feature $X = \{x_{n}\}_{n=1}^{N} \in \mathbb{R}^{N \times C}$  with $x_{n} \in \mathbb{R}^{1 \times C}$ from the back-bone network, we first project the original feature onto the feature interaction space, generating a projected feature $X^{PROJ} = \{x_{n}^{proj}\}_{n=1}^{N} \in \mathbb{R}^{K \times C'}$. $K$ is the number of the interaction nodes in the interaction graph and $C'$ is the interaction space dimension. A naive method for producing each element $x_{n}^{proj} \in X^{PROJ}, n=\{1,...,K\}$ uses the linear combination of its neighbor elements \cite{zhu2018deformable}:
\begin{equation}
x_{n}^{proj} = \sum_{\forall m \in \mathcal{N}_{n}} w_{nm}x_{m} A[n,m]
\end{equation}
where $\mathcal{N}_{n}$ denotes the neighbors of voxel $n$. The naive approach normally employs a fully-connected graph with redundant connections and parameters between the interaction nodes, being very difficult to optimise. More importantly, the linear combination method lacks an ability to perform adaptive sampling because different images contain different contextual information of brain glioma (e.g. location, size and shape). We deal with this issue by adopting the adaptive sampling strategy \cite{dai2017deformable}:
\begin{equation}
    \begin{split}
        \triangle \textit{m} &= w_{n,m}x_{n} + b_{n,m}\\
        x_{n}^{proj} &= \sum_{\forall m \in \mathcal{N}_{n}} w_{nm}\rho (x_{m} | \mathcal{V}, m, \triangle m) A[n,m]
    \end{split}
\end{equation}

where $w_{n,m} \in \mathbb{R}^{3 \times (K \times C)}$ and $b_{n,m} \in \mathbb{R}^{3 \times 1}$ are the shift distances which are learned individually for each raw feature $x_{n}$ through stochastic gradient decent. $\rho(\dot)$ is the trilinear interpolation sampler which samples a shifted feature node around feature node $x_{m}$, given the learned deformation $\triangle m$ and the total set of interaction graph nodes $\mathcal{V}$.

\textbf{Interaction Graph Reasoning} After projected the input features onto the interaction graph $G$ with $K$ feature nodes $ V = \{v_{1},...,v_{k}\}$ and edges $E$, we follow the definition of the graph convolution network \cite{DBLP:conf/iclr/KipfW17, DBLP:conf/bmvc/ZhangLAYTT19}. In particular, we define $A^{G}$ as the graph adjacency matrix on $K \times K$ nodes and $W^{G} \in \mathbb{R}^{D \times D}$ as the weight matrix, and the formulation of the graph convolution operation is formulated as follows:
\begin{equation}
\begin{split}
    X^{G} &= \sigma(A^{G}X^{PROJ} W^{G})\\
    &= \sigma((I - \hat{A^{G}}) X^{PROJ} W^{G})
\end{split}
\end{equation}
where $\sigma()$ is sigmoid activation function. We first apply Laplacian smoothing and update the adjacency matrix to $(I - \hat{A^{G}})$ so as to propagate the node feature over the entire graph.
In practice, we implement $\hat{A^{G}}$ and $W^{G}$ using a $1 \times 1$ convolution layer. We also implement $I$ as a residual connection which maximises the gradient flow \cite{he2016deep}.

\textbf{Re-Projection} Once the reasoning has been finished, we re-project the features back to the original coordinate space with output $ X^{ G} \in \mathbb{R}^{N \times D}$. We use trilinear interpolation here to calculate each graph feature instance $x^{ g}_{n} \in  X^{ G}, n \in \{1,...,N\}$ after having transformed the features from the interaction space to the coordinate space. As a result, we have the interaction graph feature $ X^{ G}$ with dimension $D$ over $N$ feature instances, identical to $X^C$.

\begin{figure*}[t]
    \centering
    \includegraphics[width=\textwidth]{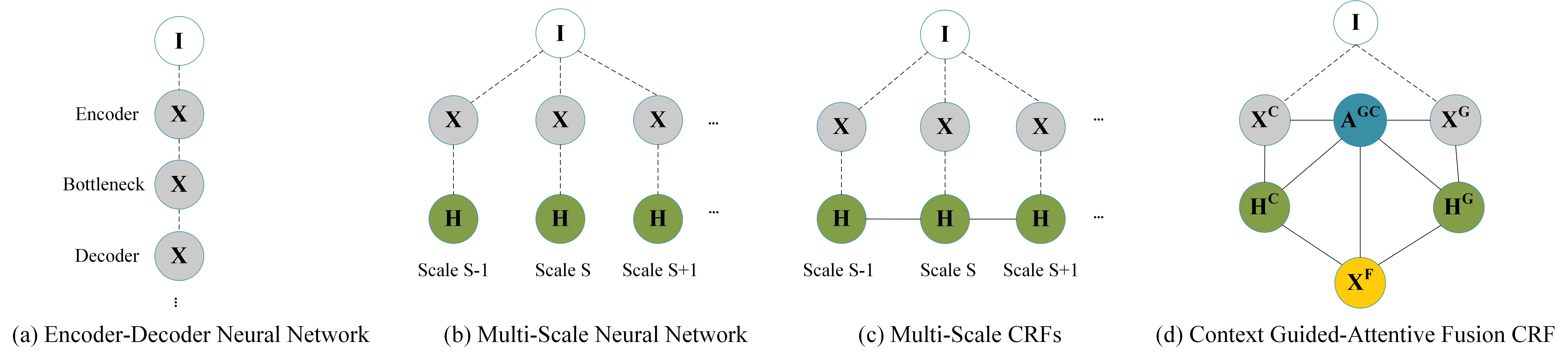}
    \caption{Graph model illustration of previous feature fusion schemes: (a) Basic encoder-decoder neural network, (b) multi-scale neural network, (c) multi-scale CRF, and (d) our proposed context guided attentive CRF. $I$ denotes the input 3D MRI image. $S$ denotes a particular feature scale. $X^C$ and $X^G$ represent the hidden features generated from the convolutional operation and graph convolutional practice respectively. $A^{GC}$ indicates the attention map generated from the corresponding feature $X^C$ and $X^G$. Best viewed in color.}
    \label{fig:crfcomp}
\end{figure*}

\subsubsection{Convolution Context Branch}
The convolution context branch is composed of an encoder and a decoder with skip connections between these two components. The encoder reduces the spatial dimensionality of the feature map whilst the expansive path recovers the feature map's spatial dimensionality and the details of objects. One of the advantages of using this architecture is that it fully exploits the features with different scales of contextual information, where large scale features can be used to localise objects and small scale yet high dimensionality features can provide more detailed and accurate information for classification.

However, networks with 3D kernels contain more parameters to learn during feature extraction. It has been observed that training such 3D model often fails in various reasons such as over-fitting and gradient vanishing or exploding. In order to address the issues mentioned above, we deploy a deep supervised mechanism for better training the convolution context branch \cite{wang2015training}. The proposed deep supervision mechanism thus reinforces the gradient flow and improves the discriminative capability during the training procedure.

Specifically, we use additional upsampling layers to reshape the features created at the deep supervised layer with the resolution of the final output. For each transform layer, we apply the softmax function to obtain additional dense segmentation maps. For these additional segmentation results, we calculate the segmentation errors with regards to the ground-truth segmentation maps. The auxiliary losses are combined with the loss from the output layer of the whole network and we further back-propagate the gradient for parameter updating during each iteration in the training stage.

We denote the set of the parameters in the deep supervised layers as $W^{S} = \{w^{s}\}_{s=1}^{S}$ and $w^{s}$ as the parameters of the upsampling layer $s$. The auxiliary loss for a deep supervision layer $s$ is formulated using cross-entropy:
\begin{equation}
    L^{s} = \sum_{n=1}^{N}-\log  \mathbbm{1}(p(y_{n}|o_{n}^{s};w^{s}))
\end{equation}
where $\mathbbm{1}$ is the indicator function which is 1 if the segmentation result is correct, otherwise 0.  $ Y = \{y_{n}\}_{n=1}^{N}$ is the ground-truth of voxel $n$ and $ O = \{o_{n}\}_{n=1}^{N}$ is the predicted segmentation label of voxel $n$ generated from the upsampling layer $s$. Finally, the deep supervision loss $ L_{s}$ can be integrated with the loss $ L^{T}$ from the final output layer. The parameters of the deep supervised layers $W^{S}$ can be updated with the rest parameters $W^{T-S}$ from the whole framework simultaneously using back-propagation:
\begin{equation}
\begin{split}
L =  L^{T}( Y| &  O;W^{T-S}, W^{S}) + \sum_{s=1}^{S}\delta^{s} L^{s}( O;w^{s})\\
      &+ \lambda(||W^{T-S}||^{2} + \sum_{s=1}^{S}(||w^{s}||^{2}))
\end{split}
\label{cnnstreamtogetherloss}
\end{equation}
where $\delta^{s}$ represents the weight for the supervision loss of each upsampling layer. As the training procedure continues to approach to the optimal parameter sets, $\delta^{s}$ reduces gradually. The final operation of Eq. (\ref{cnnstreamtogetherloss}) is the $L$2-regularisation of the total trainable weights with scalar $\lambda$.

\subsection{Context Guided Attentive CRF Fusion Module}
\label{CGA-CRF}
We further propose a novel context guided attentive CRF module to perform feature fusion, motivated from two perspectives. The graph model of our proposed CGA-CRF is illustrated in Fig. \ref{fig:crfcomp}. There are two reasons to use CGA-CRF for feature fusion. Firstly, assigning segmentation labels by maximising probabilities may result in incorrect boundary segmentation due to the neighboring voxels of sharing similar feature representation. Secondly, previous works fuse features from different sources by using a channel-wise concatenation or element-wise summation mechanism. However, these mechanisms simplifies the relationship between different source feature maps, which may result in information loss. Different from previous related works and using the inference ability of a probabilistic graphical model, we employ the conditional random field model to learn optimised latent fusion features for final settlement. As information from different contexts may contribute to the final results with different degrees, we integrate the attention gates of the CGA-CRF to regulate how much information should flow between features. We further show the implementation of CGA-CRF mean-field updates with sequential convolution operations, which allows our CGA-CRF fusion module can be integrated with any neural networks as sequential layers and trained in an end-to-end fashion. Compared with previous architectures such as encoder-decoder neural network (Fig. \ref{fig:crfcomp} (a)) and multi-scale neural network (Fig. \ref{fig:crfcomp} (b)), our proposed CGA-CRF (Fig. \ref{fig:crfcomp} (d)) has a strong inference ability and can jointly learn the hidden representation of features encoded by the neural network backbone, improving the generalisation ability of the segmentation model. Compared with previous architectures such as multi-scale CRF (Fig. \ref{fig:crfcomp} (c)), our proposed CGA-CRF model first uses an attention gate by directly modeling the cost energy in the network (Eq. (\ref{EnergyDefinition})). The attention gate thus regulates the information flow from the features encoded by the backbone neural network to the latent representations by minimising the total energy cost. We evaluate the effectiveness of each component in the experiment section.

\subsubsection{Definition}
Given the feature map $ X^{ C} = \{x^{ c}_{n}\}_{n=1}^{N}$ from the convolution context branch and the feature map $ X^{ G} = \{x^{g}_{n}\}_{n=1}^{N}$ from the interaction graph branch, our goal is to estimate the relationship between the hidden representation $H^{ G} = \{h^{ g}_{n}\}_{n=1}^{N}, H^{ C} = \{h^{ c}_{n}\}_{n=1}^{N}$, the attention variable $A^{GC} = \{a^{gc}_{n}\}_{n=1}^{N}$ and the final fused representation $ X^{ F} = \{x^{f}_{n}\}_{n=1}^{N}$. We formalise the problem by designing CGA-CRF with a gibbs distribution:
\begin{equation}
    P(X^{F},A^{GC}|I,\Theta) = \frac{1}{Z(I, \Theta)}exp\{-E(X^{F},A^{GC},I,\Theta)\}
\end{equation}
where $E(X^{F},A^{GC},I,\Theta)$ is the associated energy:
\begin{equation}
\begin{split}
     E(X^{F},A^{GC},I,\Theta) = & \Phi^{  G}(H^{  G}, X^{  G}) + \Phi^{  C}(H^{  C}, X^{  C})\\ &+ \Psi^{  G  C}(H^{  G}, H^{  C}, A^{GC})
\label{EnergyDefinition}
\end{split}
\end{equation}
where $I$ is the input 3D MRI image and $\Theta$ is the set of parameters. In Eq. (\ref{EnergyDefinition}), $\Phi^{G}$ is the unary potential between the latent graph representation $H^{G}$ and the graph features $X^{G}$. $\Phi^{C}$ is the unary potential related to latent convolution representation $H^{C}$ and convolution feature $X^{C}$. In order to drive the estimated latent representation $H$ towards the observation $X$, we use the Gaussian function created in previous works \cite{krahenbuhl2011efficient}:

\begin{equation}
    \Phi(H, X) = \sum_{n=1}^{N}\phi(h_n, x_n) = -\sum_{n=1}^{N}\frac{1}{2}||h_{n} - x_{n}||^{2}.
\end{equation}

The final term shown in Eq. (\ref{EnergyDefinition}) is the attention guided pairwise potential between the latent convolution representation $H^{C}$ and the latent graph representation $H^{G}$. The attention term $A^{GC}$ controls the information flow between the two latent representations where the graph representation may or may not contribute to the estimated convolution representation. We define:
\begin{equation}
\begin{split}
    \Psi^{GC}(H^{G}, H^{C}, & A^{GC}) = \sum_{n=1}^{N}\sum_{m \in   N_{n}}\psi(a^{gc}_{m}, h^{c}_{n}, h^{g}_{m})\\
    & = \sum_{n=1}^{N}\sum_{m \in   N_{n}}a^{gc}_{m} h^{c}_{n} \Upsilon^{GC}_{n,m} h^{g}_{m}
\end{split}
\end{equation}
The $\Upsilon^{GC}_{n,m} \in \mathbb{R}^{D^{G} \times D^{C}}$ is the kernel potential associated with hidden feature maps $H^{G}$ and $H^{C}$, where $D^{G}, D^{C}$ represents the dimensionality of the features $X^{  G}$ and $X^{  C}$ respectively.

\subsubsection{Inference}
\label{MFCRF}
By learning latent feature representations to minimise the total segmentation energy, the system can produce an appropriate segmentation map, \textit{e.g.} maximum a posterior $P(X^{F},A^{GC}|I,\Theta)$. However, the optimisation of $P(X^{F}, A^{GC}|I,\Theta)$ is intractable due to the computational complexity of the normalisation constant $Z(I, \Theta)$, which is exponentially proportional to the cardinality of $X^{F}$ and $A^{GC}$. Therefore, in order to derive the maximum a posterior in an efficient way, we adopt mean-field updates to approximate a complex posterior probability distribution:
\begin{equation}
\begin{split}
    P(X^{F},A^{GC} | I,\Theta) \approx & Q(X^{F},A^{GC}) \approx Q(H^{G}, H^{C}, A^{GC}) \\
    &= \prod_{n=1}^{N} q_{n}(h^{G}_n)q_{n}(h^{C}_n)q_{n}(a^{gc}_n)
\label{mfdef}
\end{split}
\end{equation}

Here, we use the product of independent marginal distributions $q(h^{g})$, $q(h^{c})$ and $q(a^{gc})$ to approximate the complex distribution $P(X^{F},A^{GC},I,\Theta)$. To achieve a satisfactory approximation, we minimise the Kullback-Leibler (KL) divergence $D_{KL}(Q||P)$ between the two distributions $Q$ and $P$. By replacing the definition of the energy $E(X^{F},A^{GC},I,\Theta)$, we formulate the KL divergence in Eq. (\ref{mfdef}) as follows:
\begin{equation}
\begin{split}
    D_{KL}(Q||P) &= \sum_{h}Q(h)\ln (\frac{Q(h)}{P(h)})\\
    &=\sum_{h}Q(h)E(h) + \sum_{h}Q(h)\ln Q(h) + \ln Z
\end{split}
\label{kl1}
\end{equation}
From Eq. (\ref{kl1}), we minimise the KL divergence by directly minimising the free energy $FE(Q) = \sum_{h}Q(h)E(h) + \sum_{h}Q(h)\ln (Q(h))$. In $FE(Q)$, the first item represents the cost for labelling each voxel and the second item represents the entropy of distribution $Q$. We can further expand the expression of $FE(Q)$ by replacing $Q$ and $E$ with Eqs. (\ref{mfdef}) and (\ref{EnergyDefinition}) respectively:

\begin{equation}
\begin{split}
    FE&(Q) = \sum_{n=1}^{N}q_{n}(h^{g}_n)q_{n}(h^{c}_n)q_{n}(a^{gc}_n)(\Phi^{G}+\Phi^{C}+\Psi^{GC})\\
    &+ \sum_{n=1}^{N}q_{n}(h^{g}_n)q_{n}(h^{c}_n)q_{n}(a^{gc}_n)(\ln (q_{n}(h^{g}_n)q_{n}(h^{c}_n)q_{n}(a^{gc}_n)))
\end{split}
\label{FEDef}
\end{equation}

Eq. (\ref{FEDef}) shows that the problem of minimising $FE(Q)$ can be transferred to a constrained optimisation problem with multiple variables, formulated below:

\begin{equation}
\begin{split}
    & \min_{q_{n}(h^{g}_n), q_{n}(h^{c}_n), q_{n}(a^{gc}_n)} FE(Q), \forall n \in N\\
    & \textrm{s.t.} \sum_{n=1}^{N}q_{n}(h^{g}_n) = 1, \sum_{n=1}^{N}q_{n}(h^{c}_n) = 1, \int_{0}^{1} q_{n}(a^{gc}_n)d{a^{gc}_{n}} = 1
\end{split}
\label{optDef}
\end{equation}
We can calculate the first order partial derivative by differentiating $FE(Q)$ w.r.t each variable. For example, we have:

\begin{equation}
\begin{split}
    \frac{\partial FE}{\partial q_{n}(h^{c}_{n})} = \phi^{c}(h^{c}_{n}, x^{c}_{n}) +& \sum_{m\in \mathcal{N}_{n}}\mathbb{E}_{q_{m}(a^{gc}_{m})} \{ a^{gc}_{m} \} \mathbb{E}_{q_{m}(h^{g}_n)} \psi^{gc}(h^{c}_n, h^{g}_m)\\
    &- \ln q_{n}(h^{c}_n) + \text{const}
\end{split}
\label{partialDer}
\end{equation}

By assigning 0 to the left hand side of Eq. (\ref{partialDer}), we reach:
\begin{equation}
\begin{split}
     q_{n}(h^{c}_{n}) \propto & \exp \{\phi^{c}(h^{c}_{n}, x^{c}_{n}) + \\
     & \sum_{m \in \mathcal{N}_{n}} \mathbb{E}_{q_{m}(a^{gc}_{m})} \{ a^{gc}_{m} \} \mathbb{E}_{q_{m}(h^{g}_m)} \psi (h^{c}_n, h^{g}_m)\}
\end{split}
\label{qhcdef}
\end{equation}

Eq. (\ref{qhcdef}) shows that, once the other two independent variables $q(h^{g})$ and  $q(a^{gc})$ are fixed, how $ q(h^{c})$ is updated during the mean-field approximation. Furthermore, we follow the above procedure and obtain the updating of the remaining two variable as follows:
\begin{equation}
\begin{split}
     q_{n}(h^{g}_{n}) \propto & \exp \{ \phi^{g}(h^{g}_{n}, x^{c}_{n}) + \\
     & \mathbb{E}_{q_{m}(a^{gc}_{m})} \{ a^{gc}_{m} \} \sum_{m \in \mathcal{N}_{i}} \mathbb{E}_{q_{m}(h^{c}_m)} \psi (h^{c}_n, h^{g}_m) \}
\end{split}
\label{qhgdef}
\end{equation}
\begin{equation}
    q_{n}(a^{gc}_{n}) \propto \exp \{a^{gc}_{n} \mathbb{E}_{q_{n}(h^{c}_{n})} \{ \sum_{m \in \mathcal{N}_{n}} \mathbb{E}_{q_{m}(h^{g}_{m})} \{ \psi(h^{c}_{n}, h^{g}_m)\} \} \}
\label{qatdef}
\end{equation}
where $\mathbb{E}_{q()}$ represents the expectation with respect to the distribution $q()$. Eqs. (\ref{qhcdef}-\ref{qatdef}) shown above denote the computational procedure of seeking an optimal posterior distributions of $h^{c}$, $h^{g}$ and $a^{gc}$ during the mean-field approximation. Intuitively, Eq. (\ref{qhcdef}) shows that, the latent convolution feature $h^{c}_{n}$ for voxel $n$ can be  used to describe the observation, referred to feature $x^{c}_{n}$. Afterwards, we use the re-weighted messages from the latent features of the neighboring voxels to learn the co-occurrent relationship of the pixels. The attention balance between the latent convolution and the graph features for voxel $n$ allows us to re-weight the pairwise potential message from the neighbours of voxel $n$, and then use the attention variable to re-weight the total value of voxel $n$. By denoting $\Bar{a}^{gc}_{n} = \mathbb{E}_{q(a^{gc}_{n})}\{a^{gc}_{n}\}$ and $\Bar{h}_{n} = \mathbb{E}_{q(h_{n})}\{h_{n}\}$, we have the feature update as follows:
\begin{equation}
    \Bar{h}^{g}_{n} = x^{g}_{n} + \Bar{a}^{gc}_{n} \sum_{m \in \mathcal{N}_{n}} \Upsilon^{GC}_{n,m} \Bar{h}^{c}_{m}
\end{equation}

\begin{equation}
    \Bar{h}^{c}_{n} = x^{c}_{n} + \sum_{m \in \mathcal{N}_{n}} \Bar{a}^{gc}_{m}  \Upsilon^{GC}_{n,m} \Bar{h}^{g}_{m}
\end{equation}

$\Bar{a}^{gc}_{n}$ is also derived from the probabilistic distribution, \textit{i.e.} its value lies in $[0,1]$. Here, we choose the Sigmoid function to formulate the updates for $\Bar{a}^{gc}_{n}$:
\begin{equation}
    \Bar{a}^{gc}_{n} = \sigma(- \sum_{m \in \mathcal{N}_{n}}a^{gc}_{m} h^{c}_{n} \Upsilon^{GC}_{n,m} h^{g}_{m})
\label{eqn:attenmf}
\end{equation}
where $\sigma(.)$ denotes the sigmoid activation function.

\subsubsection{CGA-CRF Inference as Convolutional Operations}

\begin{figure}
    \centering
    \includegraphics[width=0.4\textwidth]{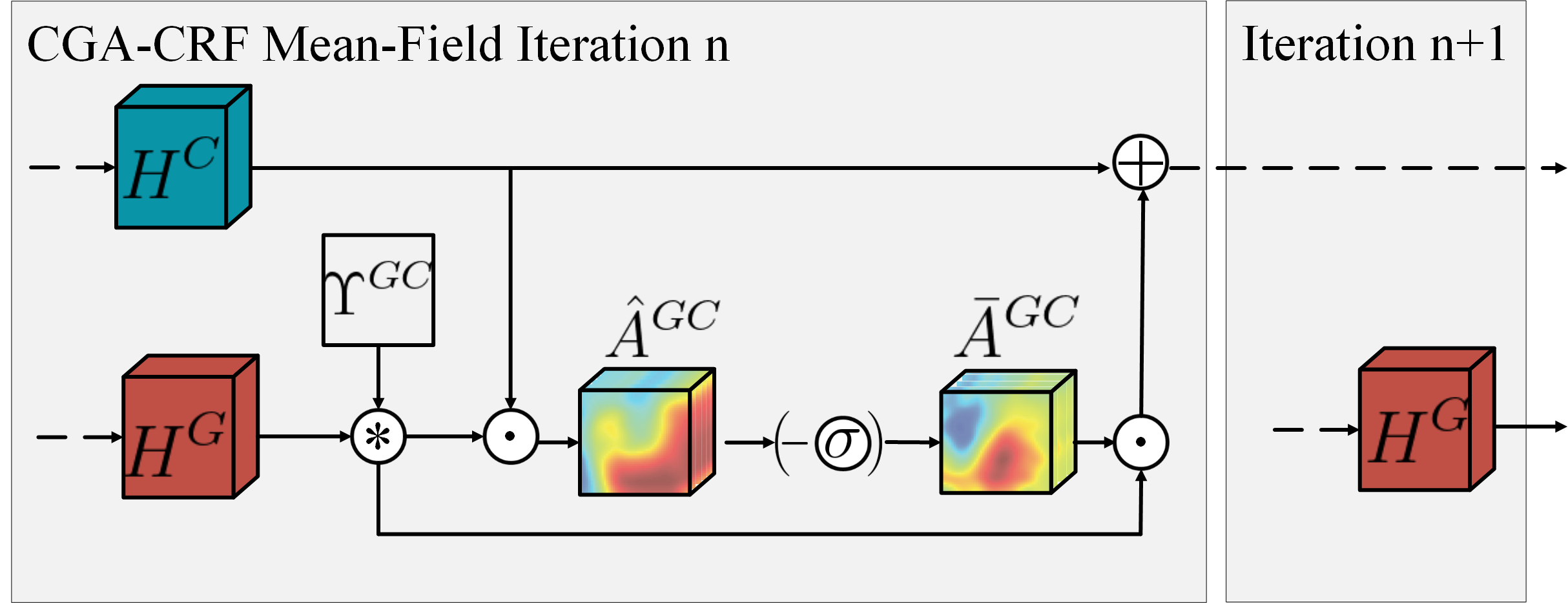}
    \caption{Details of the mean-field updates within CGA-CRF. The circled symbols indicate message-passing operations within the CGA-CRF block. Best viewed in colors.}
    \label{fig:crfmf}
\end{figure}

\begin{algorithm}
\caption{Algorithm for Mean-Field Approximation of CGA-CRF.}
\begin{algorithmic}[1]
\renewcommand{\algorithmicrequire}{\textbf{Input:}}
\renewcommand{\algorithmicensure}{\textbf{Output:}}
\REQUIRE Tensor instance from the feature interaction graph output $X^{G}$ and convolution output $X^{C}$. Initialise hidden graph feature map instance $H^{G}$ with $X^{G}$. Initialise hidden convolutional feature map instance $X^{C}$ with $X^{C}$
\ENSURE Estimated optimised feature map $H^{C}$ as $X^{F}$.
\WHILE {in iteration number}
 \STATE $\hat{A}^{GC} \leftarrow H^{C} \odot (\Upsilon^{GC} \ast H^{G})$;
 \STATE $\Bar{A}^{GC} \leftarrow \sigma(-(\hat{A}^{GC}))$;
 \STATE $H^{G} \leftarrow \Upsilon^{GC} \ast H^{G}$;
 \STATE $\Bar{H}^{C} \leftarrow \Bar{A}^{GC} \odot H^{G}$;
 \STATE $H^{C} \leftarrow X^C \oplus \Bar{H}^{C}$;
\ENDWHILE
\RETURN Optimised feature map $H^{C}$ as $X^{F}$.
\end{algorithmic}
\label{Alg:convmf}
\end{algorithm}

We implement the mean-field updates of CGA-CRF as sequential convolutional operations in order that the CGA-CRF can be trained with any neural network in an end-to-end fashion. The algorithm for implementing mean-field approximation using convolutional operations is described in Algorithm \ref{Alg:convmf}. Following Algorithm \ref{Alg:convmf}, we compile each iteration of the mean-field updates by a set of convolution operations. The output of the previous sequential convolution blocks are sent to the next sequential convolution blocks to complete one iteration. A graph illustration of Algorithm \ref{Alg:convmf} is shown in Fig. \ref{fig:crfmf}. Through this implementation, we aim to jointly estimate the hidden feature maps $H^G$, $H^C$ and the attention map $A^{GC}$ based on the derivation shown in Section \ref{MFCRF} in a data-driven manner.

Based on Eq. (\ref{eqn:attenmf}), we implement the update process of attention map $A^{GC}$ as follows: (1) Execute the message passing between hidden feature maps $H^G$ and $H^C$, $\hat{A}^{GC} \leftarrow H^{C} \odot (\Upsilon^{GC} \ast H^{G})$. Here, $\Upsilon^{GC}$ is a convolution kernel sliding on $H^G$. We set the kernel size of $\Upsilon^{GC}$ as 3. $\ast$ and $\odot$ represents the convolutional and element-wise product respectively. (2) Normalise the $\hat{A}^{GC}$ using sigmoid function $\Bar{A}^{GC} \leftarrow \sigma(-(\hat{A}^{GC}))$.

We utilise the attention map $A^{GC}$ as a switch to regulate the message flow when updating $H^G$ and $H^C$. Specifically, the mean-filed update of $H^G$ and $H^C$ can be implemented as follows: (1) Execute the message passing on $H^G$: $H^{G} \leftarrow \Upsilon^{GC} \ast H^{G}$. (2) Multiply the interaction graph feature with attention map $\Bar{H}^{C} \leftarrow \Bar{A}^{GC} \odot H^{G}$. (3) Update $H^C$ by adding the unary potential using residual connections: $H^{C} \leftarrow X^{C} \oplus \Bar{H}^{C}$, where $\oplus$ represents the element-wise summation.

By implementing the mean-field updates as sequential convolution operations, $H^G$, $H^C$ and $A^{GC}$ can be learnt and updated jointly. Note that the generalised mean field approximation is guaranteed to converge to a local optimum rather than a global optimum. Thus, to reduce the computational time of our proposed CGA-CRF, we have examined the iteration numbers of mean-field approximation of our CGA-CRF progressively and set the iteration number as 5 to reach a trade-off between competitive performance and small parameter size.

\section{Experimental Setup}
\label{Experimental Setup}
To demonstrate the effectiveness of the proposed CANet for brain glioma segmentation, we conduct experiments on three publicly available datasets: the Multimodal Brain Tumor Segmentation Challenge 2017 (BraTS2017), the Multimodal Brain Tumor Segmentation Challenge 2018 (BraTS2018) and the Multimodal Brain Tumor Segmentation Challenge 2019 (BraTS2019) \cite{menze2014multimodal, bakas2017advancing, bakas2018identifying}.  Supplementary C presents data augmentation and implementation settings.

\textbf{Datasets.} The \textbf{BraTS2017}\footnote{https://www.med.upenn.edu/sbia/brats2017.html} consists of 285 cases of patients in the training set and 44 cases in the validation set. \textbf{BraTS2018}\footnote{https://www.med.upenn.edu/sbia/brats2018.html} shares the same training set with BraTS2017 and includes 66 cases in the validation set. \textbf{BraTS2019}\footnote{https://www.med.upenn.edu/cbica/brats-2019/} expands the training set to 335 cases and the validation set to 125 cases. Each case is composed of four MR sequences, namely native T1-weighted (T1), post-contrast T1-weighted (T1ce), T2-weighted (T2) and Fluid Attenuated Inversion Recovery (FLAIR). Each sequence has a 3D MRI volume of 240$\times$240$\times$155. Ground-truth annotation is only provided in the training set, which contains the background and healthy tissues (label 0), necrotic and non-enhancing tumor (label 1), peritumoral edema (label 2) and GD-enhancing tumor (label 4). We first consider the 5-fold cross-validation on the training set where each fold contains (by random division) 228 cases for training and 57 cases for validation. We then evaluate the performance of the proposed method on the validation set. The validation result is generated from the official server of the contest to determine the segmentation accuracy of the proposed methods.

\begin{table*}[h]
\centering
\caption{Quantitative results of the CANet components by five fold cross-validation for the BraTS2017 training set (dice, sensitivity and specificity). All the methods are based on CANet with UNet as the backbone. The best result is shown in bold text and the runner-up result is underlined.}

\begin{tabular}{|c|c|c|c|c|c|c|c|c|c|c|c|c|}
\hline
              & \multicolumn{3}{c|}{DICE}                              & \multicolumn{3}{c|}{Sensitivity}                       & \multicolumn{3}{c|}{Specificity}                       & \multicolumn{3}{c|}{Hausdorff95}                       \\ \hline
Backbone+         & ET               & WT               & TC               & ET               & WT               & TC               & ET               & WT               & TC               & ET               & WT               & TC               \\ \hline
CC            & \textbf{0.686} & 0.875          & 0.821          & 0.857          & {\underline{0.925}}    & 0.863          & {\underline{0.997}}    & {\underline{0.991}}    & 0.996          & \textbf{6.791} & 6.886          & 7.939          \\ \hline
GC            & 0.637           & {\underline{0.894}}    & {\underline{0.822}}    & \textbf{0.977} & \textbf{0.970} & \textbf{0.944} & 0.987          & 0.987          & \textbf{0.997} & 9.899          & {\underline{6.403}}    & {\underline{5.812}}    \\ \hline
CC+GC+Concatenation         & 0.682          & 0.861          & 0.803          & {\underline{0.857}}    & 0.922          & 0.861          & 0.997          & 0.989          & 0.994          & {\underline{7.755}}    & 9.377          & 11.432         \\ \hline
CC+GC+CGA-CRF & {\underline{0.685}}    & \textbf{0.903} & \textbf{0.873} & 0.807          & 0.924          & {\underline{0.870}}    & \textbf{0.997} & \textbf{0.993} & {\underline{0.996}}    & 7.804          & \textbf{3.569} & \textbf{4.036} \\ \hline
\end{tabular}
\label{table:ablationstudy}
\end{table*}

\begin{figure*}[hbt]
    \centering
    \includegraphics[width=\textwidth]{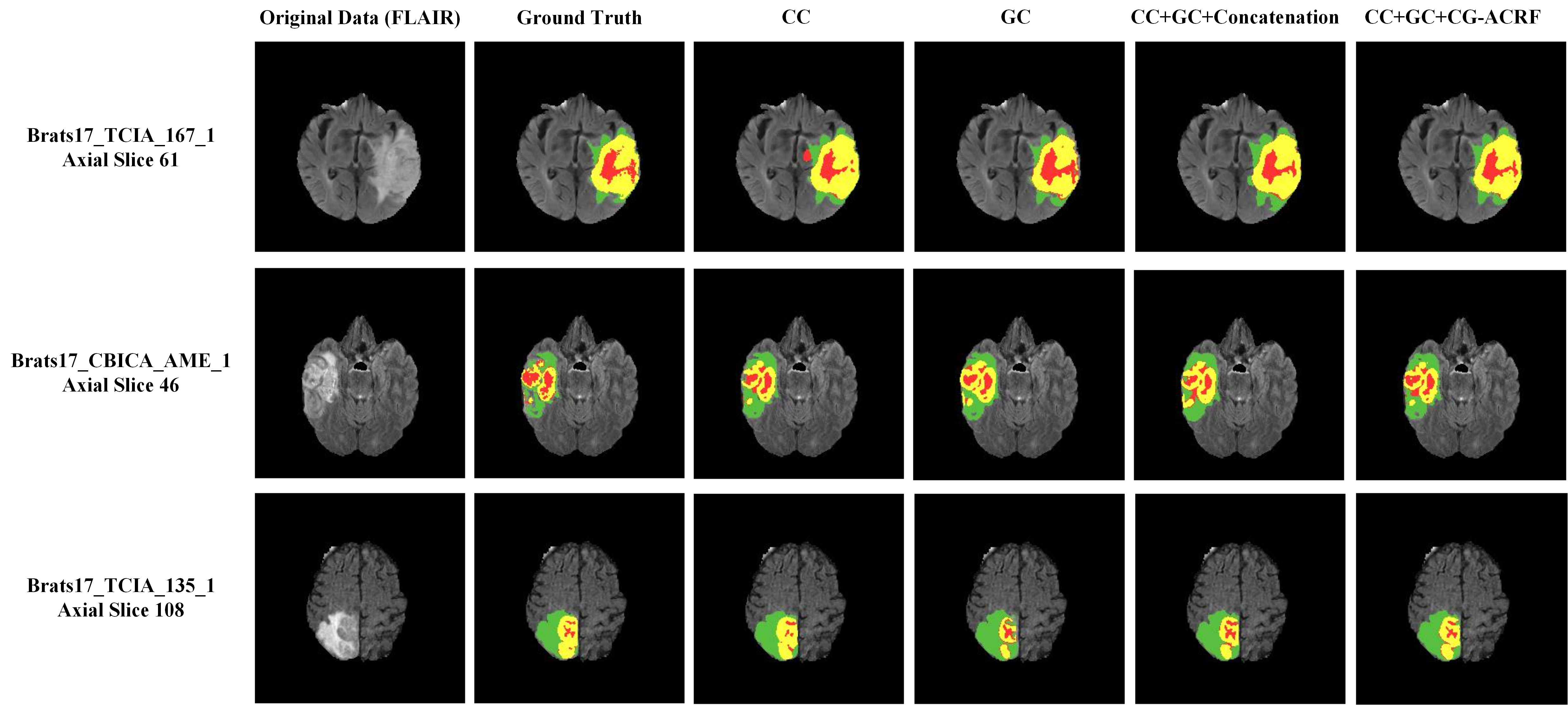}
    \caption{Qualitative comparison of different baseline models and the proposed CANet by cross validation on the BraTS2017 training set. From left to right, each column represents the input FLAIR data, ground truth annotation, segmentation result of CANet with only the convolution branch, segmentation result of CANet with only the graph convolution branch, segmentation output of CANet with HCA-FE and concatenation fusion scheme, segmentation output of CANet with HCA-FE and CG-ACRF fusion module. Best viewed in colors.}
    \label{fig:ablation}
\end{figure*}

\textbf{Evaluation Metrics.} Following previous works \cite{wang2017automatic}, \cite{kamnitsas2017efficient}, \cite{bakas2017advancing}, the segmentation accuracy is measured by Dice score, Sensitivity, Specificity and Hausdorff95 distance respectively. In particular,
\begin{itemize}
    \item Dice score: $Dice(P,T) = \frac{|P_1 \cap T_1|}{(|P_1|+|T_1|)/2}$
    \item Sensitivity: $Sens(P,T) = \frac{|P_1 \cap T_1|}{|T_1|}$
    \item Specificity: $Spec(P,T) = \frac{|P_0 \cap T_0|}{|T_0|}$
    \item Hausdorff Distance: $Haus(P,T) = \\
    max\{sup_{p\in P_1}inf_{t\in T_1} d(p,t), sup_{t\in T_1}inf_{p\in P_1} d(t,p)\}$
\end{itemize}
where $P$ represents the model prediction and $T$ represents the ground-truth annotation. $T_1$ and $T_0$ are the subset voxels predicted as positives and negatives for the tumor regions. Similar set-ups are made for $P_1$ and $P_0$. Furthermore, the Hausdorff95 measures the distance when comparing model prediction against ground-truth segmentation \cite{huttenlocher1993comparing}. $sup$ represents the supremum and $inf$ represents the infimum. For each metric, three regions namely enhancing tumor (ET, label 1), whole tumor (WT, labels 1, 2 and 4) and the tumor core (TC, labels 1 and 4) are evaluated individually.

\section{Results and Discussion}
\label{resultdiscussion}

In this section, we present both quantitative and qualitative experimental results of different methods. We first conduct an ablation study of our method to show the effective impact of building a feature interaction graph and CGA-CRF on the segmentation performance. We also perform extensive analysis on the encoder backbone and different iteration numbers of approximation in CGA-CRF. Afterwards, we compare our approach with several State-of-The-Art methods on different datasets. Finally, we present the analysis of failure cases.

\subsection{Ablation Studies} We first evaluate the impact of building the feature interaction graph and CGA-CRF. To this end, we apply 5-fold cross-evaluation on the BraTS2017 training set and report the mean results. Table \ref{table:ablationstudy} shows the quantitative results, while the qualitative results can be found in Fig. \ref{fig:ablation} as an example of the segmentation outputs. We start from two baselines. The first baseline is the fully convolution network with deep supervision on the backbone convolution encoder (CC). The second baseline only uses the proposed interaction graph in the convolution encoder without deep supervision (GC). We then evaluate the proposed (whole) CANet system (CC+GC) with concatenated feature maps from CC and GC together without any additional feature fusion method. Finally, we evaluate the proposed feature fusion module CGA-CRF, which takes the feature map with different contexts and outputs the optimal latent feature map for the final segmentation. For the experiments shown in Table \ref{table:ablationstudy} and Fig. \ref{fig:ablation}, we use the encoder of UNet as the backbone network with 5 iterations in our CGA-CRF. The experiments described later include the analysis on different backbones with iteration numbers.

\begin{table*}[hbt]
\centering
\caption{Quantitative results for different iteration numbers by CG-ACRF mean-field approximation on the five fold cross-validation of the BraTS2017 training set with respect to Dice, Sensitivity, Specificity and Hausdorff95. The best result is in bold and the runner-up result is underlined.}
\begin{tabular}{|c|c|c|c|c|c|c|c|c|c|c|c|c|}
\hline
             & \multicolumn{3}{c|}{Dice}                              & \multicolumn{3}{c|}{Sensitivity}                       & \multicolumn{3}{c|}{Specificity}                       & \multicolumn{3}{c|}{Hausdorff95}                       \\ \hline
Iteration \# & ET               & WT               & TC               & ET               & WT               & TC               & ET               & WT               & TC               & ET               & WT               & TC               \\ \hline
1            & 0.657          & 0.861          & 0.790          & \textbf{0.901} & 0.920          & 0.852          & 0.995           & 0.990          & {\underline{0.994}}    & 7.997          & 7.749          & 10.488         \\ \hline
3            & 0.681          & {\underline{0.873}}    & {\underline{0.807}}    & {\underline{0.873}}    & 0.923          & {\underline{0.869}}    & 0.996          & {\underline{0.990}}    & 0.994         & \textbf{7.614} & {\underline{6.801}}     & {\underline{8.941}}    \\ \hline
5    & \textbf{0.685} & \textbf{0.903} & \textbf{0.873} & 0.807          & {\underline{0.924}}    & \textbf{0.870} & \textbf{0.997} & \textbf{0.993} & \textbf{0.996} & {\underline{7.804}}    & \textbf{3.569} & \textbf{4.036} \\ \hline
7            & 0.664           & 0.855          & 0.769          & 0.854          & 0.921          & 0.860          & 0.996          & 0.990          & 0.993          & 9.850          & 9.720             & 12.042         \\ \hline
10           & {\underline{0.685}}    & 0.850          & 0.784           & 0.837          & \textbf{0.931} & 0.858          & {\underline{0.997}}    & 0.988          & 0.993          & 8.067          & 11.149         & 11.650         \\ \hline
\end{tabular}
\label{table:convmf}
\end{table*}

\begin{figure*}[hbt]
    \centering
    \includegraphics[width=\textwidth]{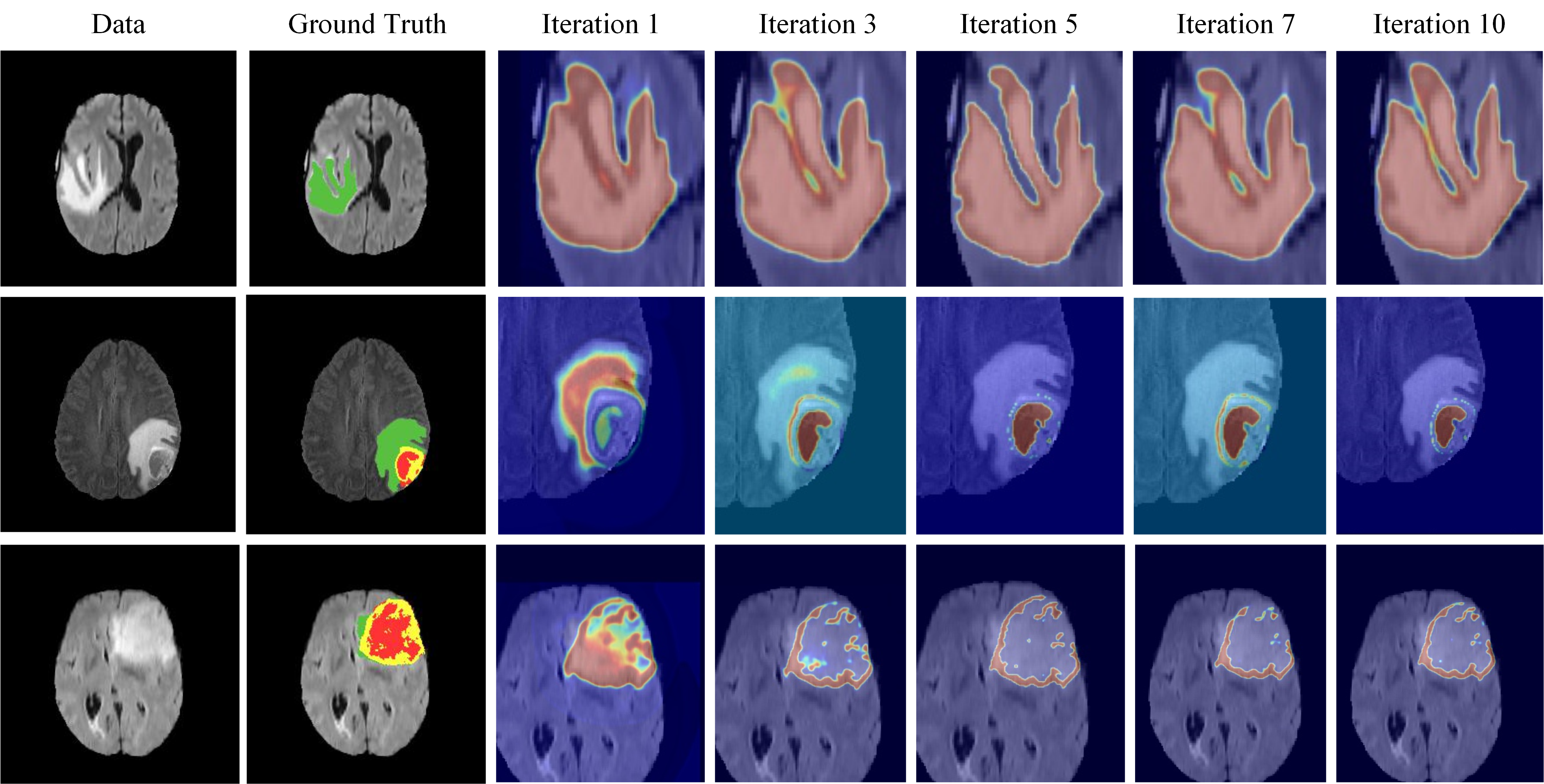}
    \caption{Examples to illustrate the results with different iteration numbers by mean-field approximation in the proposed CG-ACRF. Columns from top to bottom represent different patient cases. Rows from left to right indicate FLAIR data, ground truth annotation, attentive map generated by CANet with different iteration numbers (from 1 to 10) in CG-ACRF respectively. Best viewed in colors.}
    \label{fig:ProbMap}
\end{figure*}

From Table \ref{table:ablationstudy}, we observe that the GC obtains better performance than CC. For dice scores, GC achieves 0.894 for the entire tumor and 0.822 for the tumor core. CC only achieves a dice score of 0.875 on the entire tumor and 0.821 on the tumor core, which is 2\% and 0.2\% lower than those by GC respectively. For hausdorff95, GC achieves 6.403 on the entire tumor and 5.812 on the tumor core. CC achieves 6.886 and 7.939, which are 0.493 and 2.127 higher than those of GC on the entire tumor and the tumor core, respectively. From Fig. \ref{fig:ablation}, we observe that GC can accurately predict individual regions. For example, the GD-enhanced tumor region normally does not appear at the outside of the tumor region. This superior performance may benefit from the information learned from the feature interactive graph as the feature nodes of different tumor regions have strong structural association between them. Learning the relationship may help the system to predict correct labels of the tumor regions. However, the sensitivity of GC is much higher than that of CC. In Table \ref{table:ablationstudy}, for example, the sensitivity score of GC is higher than that of CC: 12.02\% higher on the enhancing tumor, 4.469\% higher on the entire tumor, 8.104\% higher on tumor core, respectively. We observe poor segmentation results at the NCR/ECT region by GC, inferior to CC with the ground truth shown in Fig. \ref{fig:ablation}.

\begin{table*}[h]
\centering
\caption{Quantitative results of the State-of-The-Art models by cross-validation on the BraTS2017 training set with respect to dice, sensitivity, specificity and hausdorff. The best result is shown in bold and the runner-up result is underlined.}

\begin{tabular}{|c|c|c|c|c|c|c|c|c|c|c|c|c|}
\hline
               & \multicolumn{3}{c|}{Dice}                              & \multicolumn{3}{c|}{Sensitivity}                       & \multicolumn{3}{c|}{Specificity}                       & \multicolumn{3}{c|}{Hausdorff95}                       \\ \hline
Model          & ET               & WT               & TC               & ET               & WT               & TC               & ET               & WT               & TC               & ET               & WT               & TC               \\ \hline
3D-UNet\cite{cciccek20163d}        & 0.706          & 0.865          & 0.810          & 0.803          & 0.906           & 0.829          & 0.998          & 0.990          & 0.995          & 6.624          & 8.193          & 8.958          \\ \hline
No-New Net\cite{isensee2018no}     & {\textbf{0.741}}    & 0.871          & 0.812           & 0.767          & 0.893          & 0.831          & \textbf{0.999} & {\underline{0.992}}    & 0.995          & \textbf{3.930} & 7.055          & 7.641          \\ \hline
Attention UNet\cite{oktay2018attention} & 0.672          & 0.863           & 0.778          & \textbf{0.847} & 0.900           & 0.862          & 0.996          & 0.990          & 0.992          & 9.347          & 9.676          & 10.668         \\ \hline
PRUNet\cite{brugger2019partially}         & \uline{0.710}          & 0.891          & 0.814          & 0.788          & 0.900          & 0.841          & {\underline{0.998}}    & 0.990          & 0.996          & 7.205          & 7.414          & 9.187           \\ \hline
3D-ESPNet\cite{mehta20193despnet}      & 0.690          & {\underline{0.895}}    & {\underline{0.844}}    & 0.805          & \textbf{0.947} & \textbf{0.881} & 0.997          & 0.990          & {\underline{0.997}}    & 6.894          & {\underline{4.156}}    & {\underline{5.778}}    \\ \hline
CANet (Ours)   & 0.685          & \textbf{0.903} & \textbf{0.873} & 0.807          & {\underline{0.924}}    & {\underline{0.870}}    & 0.997          & \textbf{0.993} & 0.996          & 7.804          & \textbf{3.569} & \textbf{4.036} \\ \hline
\end{tabular}
\label{table:SOTACompare}
\end{table*}

\begin{figure*}[h]
    \centering
    \includegraphics[width=\textwidth]{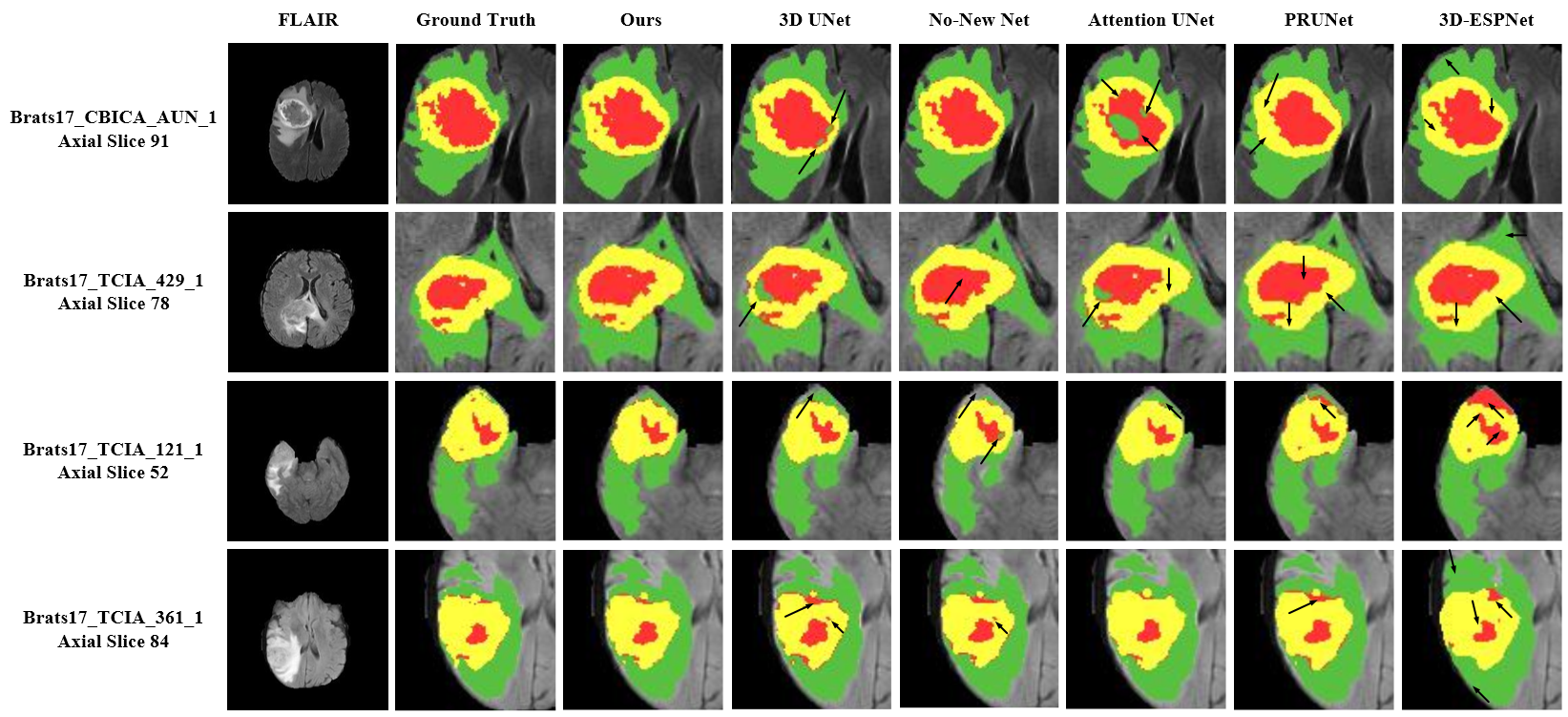}
    \caption{Examples of segmentation results by cross validation on the BraTS2017 training set. Qualitative comparisons with other brain glioma segmentation methods are presented. The eight columns from left to right show the frames of the input FLAIR data, the ground truth annotation, the results generated from our CANet (UNet encoder backbone and 5-iteration CGA-CRF), 3DUNet \cite{cciccek20163d}, NoNewNet \cite{isensee2018no}, Attention UNet \cite{oktay2018attention}, PRUNet \cite{mehta20193despnet}, respectively. Black arrows indicate the failure cases in these comparison methods. Best viewed in colors.}
    \label{fig:SOTA}
\end{figure*}

\begin{figure}[h]
    \centering
    \includegraphics[width=0.5\textwidth]{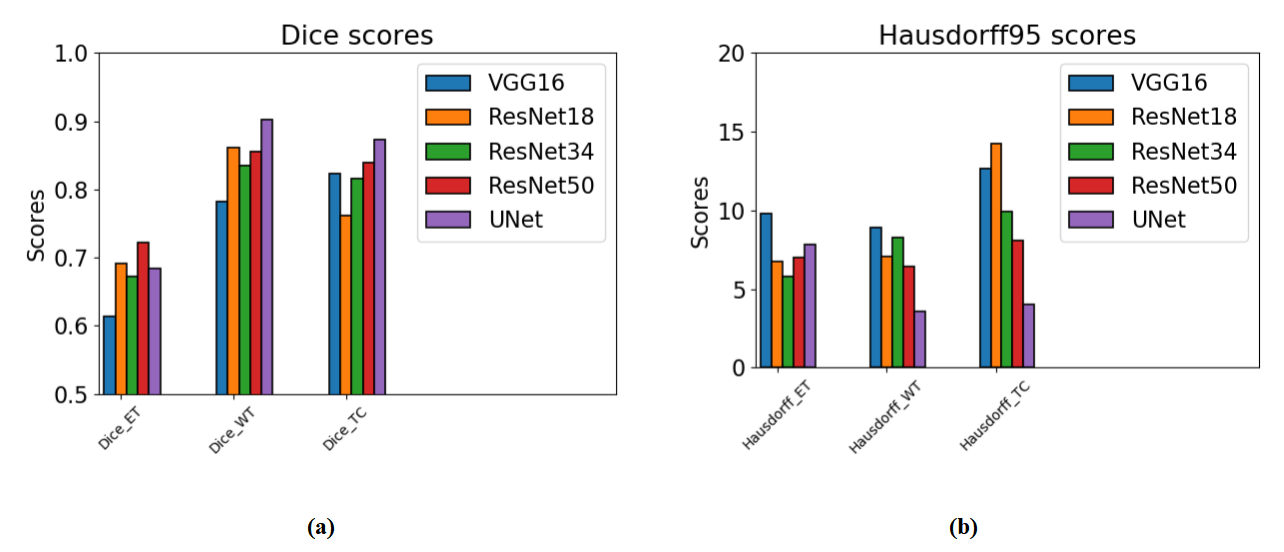}
    \caption{Performance comparison with different encoder backbones: (a) and (b) indicate the comparisons with dice score and hausdorff95 by cross validation on the BraTS2017 training set using different encoder backbones respectively. Best viewed in colors.}
    \label{fig:backbone}
\end{figure}

We then evaluate the complete CANet with the extracted feature maps by CC and GC simultaneously. Here, we fuse the feature maps of CC and GC using a naive concatenation method, which has less over-segmentation results. Depicted in Table \ref{table:ablationstudy}, the sensitivity of CC+GC is much lower than that of GC. The sensitivity of CC+GC is  0.857 on the enhancing tumor (ET), 0.922 on the whole tumor (WT) and 0.861 on the tumor core (TC), respectively. From Fig. \ref{fig:ablation}, we witness that by introducing the feature interaction graph, the segmentation model can correct some misclassified regions produced by CC. However, the concatenation fusion method does not demonstrate any benefit on the overall segmentation performance. CC+GC has a dice score of 0.861 on the whole tumor and 0.803 on the tumor core, which are 3.292\% and 1.94\% lower than those of GC respectively. We also observe the loss of the boundary information shown in Fig. \ref{fig:ablation}, especially the boundaries of NCR/ECT and GD-enhancing tumors excessively shrinks compared with those of GC and CC.

We finally evaluate the effectiveness of our proposed CGA-CRF. By introducing the CGA-CRF fusion module, our segmentation model outperforms the other methods. Benefiting from the inference ability of CGA-CRF, it presents a satisfactory segmentation output. For the whole tumor and the tumor core, its Dice scores are 0.903 and 0.873 respectively, which are the top scores in the leader-board. Its Hausdorff95 also is the lowest. For the whole tumor and the tumor core, its hausdorff95 values are 3.569 and 4.036 respectively. Referring to much lower sensitivity scores reported in Table \ref{table:ablationstudy}, we conclude that the superior performance has been achieved by the complete CANet. The same conclusion can be drawn from Fig. \ref{fig:ablation} where CGA-CRF can detect optimal feature maps that benefit the downstream deconvolution networks and outline small tumor cores and edges, which may be lost when we use a down-sampling operation in the encoder backbone.

\begin{table*}[h]
\centering
\caption{Quantitative comparisons between CANet and the other state of the art techniques on the BraTS2017 validation set with respect to Dice and Hausdorff95. The best results out of each category are shown in bold. '-' depicts that the result of the associated method has not been reported yet. \text{$\star$} and \text{$\dagger$} represents the final winner and runner-up solution respectively.}
\begin{tabular}{|cc|l|ccc|ccc|c|}
\hline
                                        &                                                                                                       &         & \multicolumn{3}{c|}{Dice}                                                                                                                     & \multicolumn{3}{c|}{Hausdorff95}                                                                                                              & \begin{tabular}[c]{@{}c@{}}Model\\ Parameter\end{tabular}\\  \hline
\multicolumn{1}{|c|}{Approach}                                & Method                                                                                                &         & ET                                            & WT                                            & TC                                            & ET                                            & WT                                            & TC                                            &                                                                                                                                   \\ \hline
\multicolumn{1}{|c|}{}                  & \multirow{2}{*}{Kamnitsas et al. \cite{kamnitsas2017ensembles} $\star$} & Mean:   & 0.738                                         & 0.901                                         & 0.797                                         & 4.500                                         & 4.230                                         & 6.560                                         & \multirow{2}{*}{-}                                                                                                                                 \\
\multicolumn{1}{|l|}{}                  &                                                                                                       & StdDev: & -                                             & -                                             & -                                             & -                                             & -                                             & -                                             & \multicolumn{1}{l|}{}                                                                                                             \\ \cline{2-10} 
\multicolumn{1}{|c|}{}                  & \multirow{2}{*}{Wang et al. \cite{wang2017automatic}  $\dagger$}                         & Mean:   & \textbf{0.786} & \textbf{0.905} & \textbf{0.838} & \textbf{3.282} & \textbf{3.890} & \textbf{6.479} & \multirow{2}{*}{5.95E5}                                                                                                                            \\
\multicolumn{1}{|l|}{}                  & \multicolumn{1}{l|}{}                                                                                 & StdDev: & -                                             & -                                             & -                                             & -                                             & -                                             & -                                             & \multicolumn{1}{l|}{}                                                                                                             \\ \cline{2-10} 
\multicolumn{1}{|c|}{Ensemble}          & \multirow{2}{*}{Zhao et al. \cite{zhao20173d}}                                                        & Mean:   & 0.754                                         & 0.887                                         & 0.794                                         & -                                             & -                                             & -                                             & \multirow{2}{*}{-}                                                                                                                                 \\
\multicolumn{1}{|l|}{}                  & \multicolumn{1}{l|}{}                                                                                 & StdDev: & -                                             & -                                             & -                                             & -                                             & -                                             & -                                             & \multicolumn{1}{l|}{}                                                                                                             \\ \cline{2-10} 
\multicolumn{1}{|c|}{}                  &\multirow{2}{*}{ Isensee et al. \cite{isensee2017brain}}                                               & Mean:   & 0.732                                         & 0.896                                         & 0.797                                         & 4.550                                         & 6.970                                         & 9.480                                         & \multirow{2}{*}{-}                                                                                                                                 \\
\multicolumn{1}{|l|}{}                  & \multicolumn{1}{l|}{}                                                                                 & StdDev: & -                                             & -                                             & -                                             & -                                             & -                                             & -                                             & \multicolumn{1}{l|}{}                                                                                                             \\ \cline{2-10} 
\multicolumn{1}{|c|}{}                  & \multirow{2}{*}{Jungo et al. \cite{jungo2017towards}}                                                 & Mean:   & 0.749                                         & 0.901                                         & 0.790                                         & 5.379                                         & 5.409                                         & 7.487                                         & \multirow{2}{*}{-}                                                                                                                                 \\
\multicolumn{1}{|l|}{}                  & \multicolumn{1}{l|}{}                                                                                 & StdDev: & 0.277                                         & 0.086                                         & 0.239                                         & 10.068                                        & 9.710                                         & 8.935                                         & \multicolumn{1}{l|}{}                                                                                                             \\ \hline
\multicolumn{1}{|c|}{}                  &\multirow{2}{*}{ Islam et al. \cite{islam2017multi}}                                                   & Mean:   & 0.689                                         & 0.876                                         & 0.761                                         & 12.938                                        & 9.820                                         & 12.361                                        & \multirow{2}{*}{1.34E8}                                                                                                                            \\
\multicolumn{1}{|l|}{}                  & \multicolumn{1}{l|}{}                                                                                 & StdDev: & 0.304                                         & 0.086                                         & 0.221                                         & 26.453                                        & 13.516                                        & 20.826                                        & \multicolumn{1}{l|}{}                                                                                                             \\ \cline{2-10} 
\multicolumn{1}{|c|}{}                  & \multirow{2}{*}{Lopez et al. \cite{lopez2017dilated}}                                                 & Mean:   & 0.567                                         & 0.783                                         & 0.685                                         & 23.828                                        & 30.316                                        & 38.077                                        & \multirow{2}{*}{1.57E7}                                                                                                                            \\
\multicolumn{1}{|l|}{}                  & \multicolumn{1}{l|}{}                                                                                 & StdDev: & -                                             & -                                             & -                                             & -                                             & -                                             & -                                             & \multicolumn{1}{l|}{}                                                                                                             \\ \cline{2-10} 
\multicolumn{1}{|c|}{}                  & \multirow{2}{*}{Shaikh et al. \cite{shaikh2017brain}}                                                 & Mean: & 0.650                                         & 0.870                                         & 0.680                                         & -                                             & -                                             & -                                             & \multirow{2}{*}{2.31E7}                                                                                                                            \\
\multicolumn{1}{|l|}{}                  & \multicolumn{1}{l|}{}                                                                                 & StdDev: & 0.320                                         & 0.110                                         & 0.340                                         & -                                             & -                                             & -                                             & \multicolumn{1}{l|}{}                                                                                                             \\ \cline{2-10} 
\multicolumn{1}{|c|}{}                  & \multirow{2}{*}{Castillo et al \cite{castillo2017volumetric}}                                         & Mean:   & 0.690                                         & 0.860                                         & 0.690                                         & -                                             & -                                             & -                                             & \multirow{2}{*}{2.22E7}                                                                                                                            \\
\multicolumn{1}{|l|}{}                  & \multicolumn{1}{l|}{}                                                                                 & StdDev: & -                                             & -                                             & -                                             & -                                             & -                                             & -                                             & \multicolumn{1}{l|}{}                                                                                                             \\ \cline{2-10} 
\multicolumn{1}{|c|}{Single Prediction} & \multirow{2}{*}{Li et al. \cite{li2017compactness}}                                                   & Mean:   & 0.704                                         & 0.871                                         & 0.682                                         & 7.699                                         & 10.396                                        & 13.062                                        & \multirow{2}{*}{3.80E5}                                                                                                                            \\
\multicolumn{1}{|l|}{}                  & \multicolumn{1}{l|}{}                                                                                 & StdDev: & 0.307                                         & 0.083                                         & 0.304                                         & 14.407                                        & 15.754                                        & 17.573                                        & \multicolumn{1}{l|}{}                                                                                                             \\ \cline{2-10} 
\multicolumn{1}{|c|}{}                  & \multirow{2}{*}{Jesson et al. \cite{jesson2017brain}}                                                 & Mean:   & 0.713                                         & \textbf{0.899}               & 0.751                                         & 6.980                                         & \textbf{4.160}               & \textbf{8.650}               & \multirow{2}{*}{4.29E6}                                                                                                                            \\
\multicolumn{1}{|l|}{}                  & \multicolumn{1}{l|}{}                                                                                 & StdDev: & 0.291                                         & 0.070                                         & 0.240                                         & 12.100                                        & 3.370                                         & 9.350                                         & \multicolumn{1}{l|}{}                                                                                                             \\ \cline{2-10} 
\multicolumn{1}{|c|}{}                  & \multirow{2}{*}{Roy et al. \cite{roy2018recalibrating}}                                               & Mean:   & 0.716                                         & 0.892                                         & 0.793                                         & 6.612                                         & 6.735                                         & 9.806                                         & \multirow{2}{*}{-}                                                                                                                                 \\
\multicolumn{1}{|l|}{}                  & \multicolumn{1}{l|}{}                                                                                 & StdDev: & -                                             & -                                             & -                                             & -                                             & -                                             & -                                             & \multicolumn{1}{l|}{}                                                                                                             \\ \cline{2-10} 
\multicolumn{1}{|c|}{}                  & \multirow{2}{*}{Pereira er al. \cite{pereira2019adaptive}}                                            & Mean:   & 0.719                                         & 0.889                                         & 0.758                                         & 5.738                                         & 6.581                                         & 11.100                                        & \multirow{2}{*}{-}                                                                                                                                \\
\multicolumn{1}{|l|}{}                  & \multicolumn{1}{l|}{}                                                                                 & StdDev: & -                                             & -                                             & -                                             & -                                             & -                                             & -                                             & \multicolumn{1}{l|}{}                                                                                                             \\ \cline{2-10} 
\multicolumn{1}{|c|}{}                  & \multirow{2}{*}{CANet (Ours)}                                                                                          & Mean:   & \textbf{0.728}               & 0.892                                         & \textbf{0.821}               & \textbf{5.496}               & 7.392                                         & 10.122                                        & \multirow{2}{*}{3.34E7}                                                                                                                            \\
\multicolumn{1}{|l|}{}                  & \multicolumn{1}{l|}{}                                                                                 & StdDev: & 0.286                                         & 0.082                                         & 0.167                                         & 11.690                                        & 11.917                                        & 16.966                                        & \multicolumn{1}{l|}{}                                                                                                             \\ \hline
\end{tabular}
\label{table:17val}
\end{table*}

\begin{figure}[h]
    \centering
    \includegraphics[width=0.48 \textwidth]{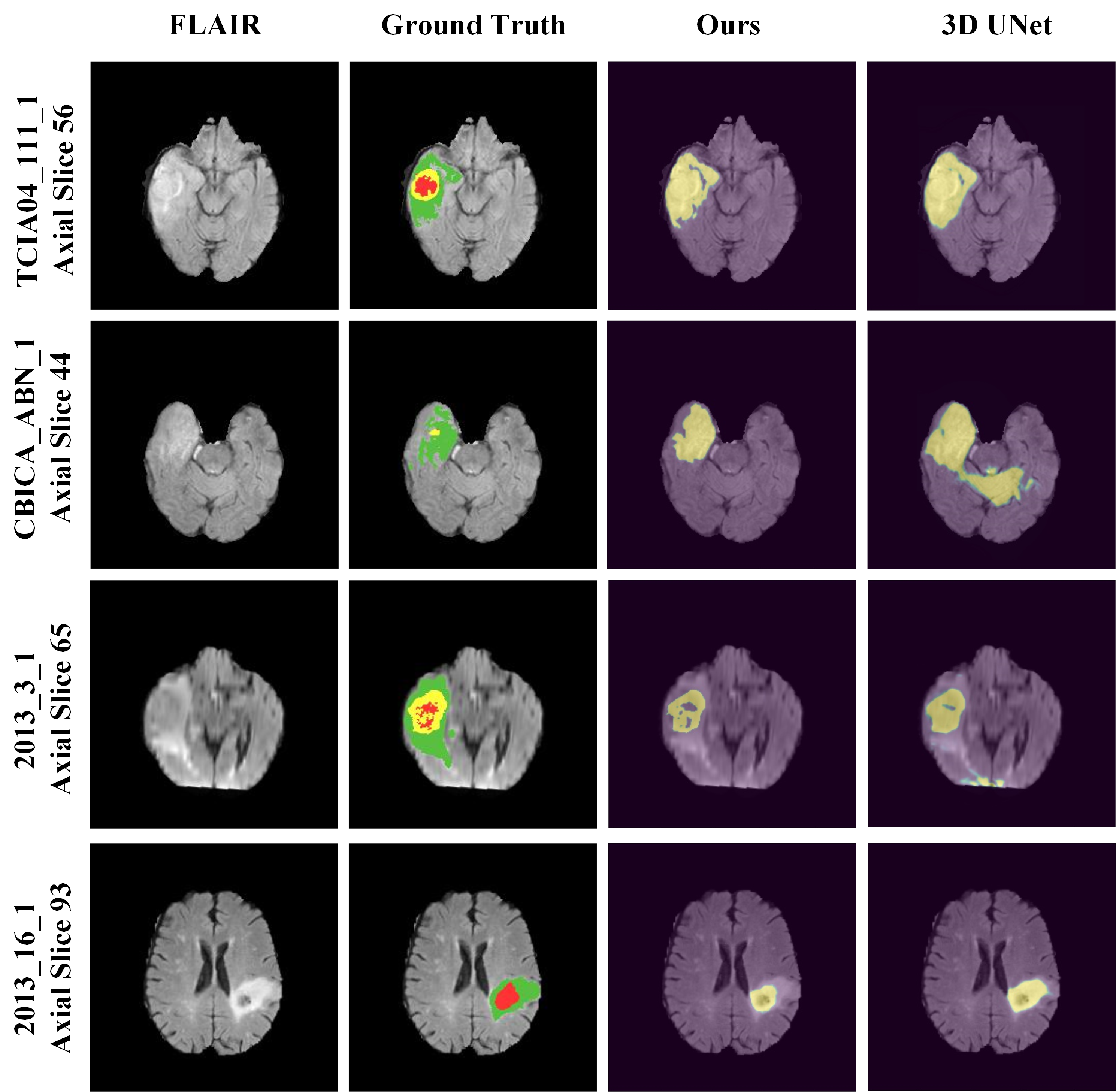}
    \caption{Examples of segmentation probability maps of our proposed CANet and 3D UNet. Columns from top to bottom represent different patient cases. Rows from left to right indicate the FLAIR data, ground truth annotation, attentive map generated by CANet with 5 iterations in CG-ACRF and attentive map generated by 3D UNet respectively. Best viewed in colors.}
    \label{fig:probComp}
\end{figure}

\begin{table*}[h]
\centering
\caption{Quantitative results of the BraTS2018 validation set with respect to Dice and Hausdorff95. The best results out of each category are shown in bold. '-' represents the result of the associated method has not been reported yet. \text{$\star$} and \text{$\dagger$} represents the final winner and the runner-up solution respectively.}
\begin{tabular}{|cc|l|ccc|ccc|c|}
\hline
                                        &                                                                      &         & \multicolumn{3}{c|}{Dice}                                                                           & \multicolumn{3}{c|}{Hausdorff95}                                                                    & \begin{tabular}[c]{@{}c@{}}Model\\ Parameter\end{tabular}\\ \hline
\multicolumn{1}{|c|}{Approach}          & Method                                                               &         & ET                              & WT                              & TC                              & ET                              & WT                              & TC                              &                                                                                                                                   \\ \hline
\multicolumn{1}{|c|}{}                  & \multirow{2}{*}{Isensee et al. \cite{isensee2018no} $\dagger$} & Mean:   & 0.796                           & \textbf{0.908} & 0.843                           & 3.120                           & 4.790                           & 8.020                           & \multirow{2}{*}{1.45E7}                                                                                                           \\
\multicolumn{1}{|l|}{}                  &                                                                      & StdDev: & -                               & -                               & -                               & -                               & -                               & -                               &                                                                                                                                   \\ \cline{2-10} 
\multicolumn{1}{|c|}{}                  & \multirow{2}{*}{McKinley et al. \cite{mckinley2018ensembles}}               & Mean:   & 0.793                           & 0.901                           & 0.847                           & 3.603                           & 4.062                           & \textbf{4.988} & \multirow{2}{*}{-}                                                                                                                \\
\multicolumn{1}{|l|}{}                  &                                                                      & StdDev: & -                               & -                               & -                               & -                               & -                               & -                               &                                                                                                                                   \\ \cline{2-10} 
\multicolumn{1}{|c|}{}                  & \multirow{2}{*}{Zhou et al. \cite{zhou2018learning}}                        & Mean:   & 0.792                           & 0.907                           & 0.836                           & 2.800                           & 4.480                           & 7.070                           & \multirow{2}{*}{-}                                                                                                                \\
\multicolumn{1}{|l|}{}                  &                                                                      & StdDev: & -                               & -                               & -                               & -                               & -                               & -                               &                                                                                                                                   \\ \cline{2-10} 
\multicolumn{1}{|c|}{Ensemble}          & \multirow{2}{*}{Cabezas et al. \cite{cabezas2018survival}}                  & Mean:   & 0.740                           & 0.889                           & 0.726                           & 5.304                           & 6.956                           & 11.924                          & \multirow{2}{*}{-}                                                                                                                \\
\multicolumn{1}{|l|}{}                  &                                                                      & StdDev: & 0.277                           & 0.075                           & 0.243                           & 9.964                           & 11.939                          & 13.448                          &                                                                                                                                   \\ \cline{2-10} 
\multicolumn{1}{|c|}{}                  & \multirow{2}{*}{Puch et al. \cite{puch2018global}}                          & Mean:   & 0.758                           & 0.895                           & 0.774                           & 4.502                           & 10.656                          & 7.103                           & \multirow{2}{*}{1.48E6}                                                                                                           \\
\multicolumn{1}{|l|}{}                  &                                                                      & StdDev: & 0.264                           & 0.070                           & 0.253                           & 8.227                           & 19.286                          & 7.084                           &                                                                                                                                   \\ \cline{2-10} 
\multicolumn{1}{|c|}{}                  & \multirow{2}{*}{Feng et al. \cite{feng2018brain}}                           & Mean:   & 0.787                           & 0.906                           & 0.834                           & 3.964                           & \textbf{4.018} & 5.340                           & \multirow{2}{*}{-}                                                                                                                \\
\multicolumn{1}{|l|}{}                  &                                                                      & StdDev: & -                               & -                               & -                               & -                               & -                               & -                               &                                                                                                                                   \\ \cline{2-10} 
\multicolumn{1}{|c|}{}                  & \multirow{2}{*}{Sun et al. \cite{sun2018tumor}}                             & Mean:   & \textbf{0.805} & 0.904                           & \textbf{0.849} & \textbf{2.777} & 6.327                           & 6.373                           & \multirow{2}{*}{-}                                                                                                                \\
\multicolumn{1}{|l|}{}                  &                                                                      & StdDev: & -                               & -                               & -                               & -                               & -                               & -                               &                                                                                                                                   \\ \hline
\multicolumn{1}{|c|}{}                  & \multirow{2}{*}{Carver et al. \cite{carver2018automatic}}                   & Mean:   & 0.710                           & 0.880                           & 0.770                           & 4.460                           & 7.090                           & 9.570                           & \multirow{2}{*}{2.20E7}                                                                                                           \\
\multicolumn{1}{|l|}{}                  &                                                                      & StdDev: & 0.290                           & 0.080                           & 0.260                           & 8.320                           & 11.570                          & 14.080                          &                                                                                                                                   \\ \cline{2-10} 
\multicolumn{1}{|c|}{}                  & \multirow{2}{*}{Chen et al. \cite{chen2018focus}}                           & Mean:   & 0.707                           & 0.845                           & 0.731                           & 10.385                          & 11.822                          & 15.066                          & \multirow{2}{*}{1.33E7}                                                                                                           \\
\multicolumn{1}{|l|}{}                  &                                                                      & StdDev: & 0.264                           & 0.100                           & 0.230                           & 21.205                          & 23.610                          & 20.560                          &                                                                                                                                   \\ \cline{2-10} 
\multicolumn{1}{|c|}{Single Prediction} & \multirow{2}{*}{Salehi et al. \cite{salehi2017auto}}                        & Mean:   & 0.704                           & 0.822                           & 0.733                           & 9.668                           & 9.610                           & 13.909                          & \multirow{2}{*}{1.20E7}                                                                                                           \\
\multicolumn{1}{|l|}{}                  &                                                                      & StdDev: & 0.289                           & 0.136                           & 0.242                           & 13.757                          & 13.036                          & 14.965                          &                                                                                                                                   \\ \cline{2-10} 
\multicolumn{1}{|c|}{}                  & \multirow{2}{*}{Myronenko \cite{myronenko20183d} $\star$}      & Mean:   & \textbf{0.816} & \textbf{0.904} & \textbf{0.860} & \textbf{3.805} & \textbf{4.483} & 8.278                           & \multirow{2}{*}{2.01E7}                                                                                                           \\
\multicolumn{1}{|l|}{}                  &                                                                      & StdDev: & -                               & -                               & -                               & -                               & -                               & -                               &                                                                                                                                   \\ \cline{2-10} 
\multicolumn{1}{|c|}{}                  & \multirow{2}{*}{Weninger et al. \cite{weninger2018segmentation}}            & Mean:   & 0.712                           & 0.889                           & 0.758                           & 8.628                           & 6.970                           & 10.910                          & \multirow{2}{*}{-}                                                                                                                \\
\multicolumn{1}{|l|}{}                  &                                                                      & StdDev: & -                               & -                               & -                               & -                               & -                               & -                               &                                                                                                                                   \\ \cline{2-10} 
\multicolumn{1}{|c|}{}                  & \multirow{2}{*}{Gates et al. \cite{gates2018glioma}}                        & Mean:   & 0.678                           & 0.806                           & 0.685                           & 14.523                          & 14.415                          & 20.017                          & \multirow{2}{*}{-}                                                                                                                \\
\multicolumn{1}{|l|}{}                  &                                                                      & StdDev: & -                               & -                               & -                               & -                               & -                               & -                               &                                                                                                                                   \\ \cline{2-10} 
\multicolumn{1}{|c|}{}                  & \multirow{2}{*}{CANet(Ours)}                                         & Mean:   & 0.767                           & 0.898                           & 0.834                           & 3.859                           & 6.685                           & \textbf{7.674} & \multirow{2}{*}{3.34E7}                                                                                                           \\
\multicolumn{1}{|l|}{}                  &                                                                      & StdDev: & 0.247                           & 0.082                           & 0.167                           & 11.690                          & 10.135                          & 14.981                          &                                                                                                                                   \\ \hline
\end{tabular}
\label{table:18val}
\end{table*}

\begin{table*}[h]
\centering
\caption{Quantitative results of the BraTS2019 validation set with respect to Dice and Hausdorff95. The best result is shown in bold and the runner-up result is underlined. \text{$\star$} represents the final winner solution.}
\begin{tabular}{|c|ccc|ccc|}
\hline
                 & \multicolumn{3}{c|}{Dice}                                     & \multicolumn{3}{c|}{Hausdorff95} \\ \hline
Method           & ET                        & WT                        & TC    & ET        & WT        & TC       \\ \hline
Jiang et al.\cite{jiang2019two}  $\star$   & \textbf{0.802}                     & \textbf{0.908}                     & \textbf{0.863} & \textbf{3.206}     & \textbf{4.444}     & \textbf{5.862}    \\
Zhao et al.\cite{zhao2019bag}      & 0.702                     & 0.893                     & 0.800 & 4.766     & \uline{5.078}     & \uline{6.472}    \\
Wang et al.\cite{wang20193d}      & 0.737                     & \uline{0.894}                     & 0.807 & 5.994     & 5.677     & 7.357    \\
Li et al.\cite{li2019multi}        & 0.771                     & 0.886                     & 0.813 & 6.033     & 6.232     & 7.409    \\
Myronenko et al.\cite{myronenko2019robust} & \uline{0.800}                     & \uline{0.894}                     & 0.834 & \uline{3.921}     & 5.890     & 6.562    \\
CANet (Ours)     & \multicolumn{1}{l}{0.759} & \multicolumn{1}{l}{0.885} & \uline{0.851} & 4.809     & 7.091     & 8.409    \\ \hline
\end{tabular}
\label{table:19val}
\end{table*}


\textbf{Backbone Test} We then evaluate the effectiveness of different encoder backbones. To do so, we use 5 fold cross-validation on the BraTS2017 training set with complete CC+GC and 5-iteration CGA-CRF. We here choose the State-of-The-Art encoder backbones, e.g. VGG16, ResNet18, ResNet30, ResNet50 and UNet encoder path. For each backbone, we feed the feature map from the last convolution block into the feature interaction graph branch to extract the interaction graph contexts and feed the feature map from the second last convolution block into the convolution branch to generate deep supervised feature maps. This practice has been proved to be effective, efficient and simple. The segmentation results with respect to Dice and Hausdorff95 are shown in Fig. \ref{fig:backbone}. ResNet outperforms the VGG16 mainly due to the involved residual connection and batch normalisation. However, comparing ResNet and the encoder of UNet, the encoder of UNet achieves better segmentation performance in terms of Dice and Hausdorff95 due to the multiple scale feature maps and skip connection for feature fusion. We choose the encoder of UNet as the backbone network for the final segmentation model in our approach.

\textbf{Iteration Test} we manually set the iteration number in the mean-field approximation of CG-ACRF. Since the mean-field approximation can only guarantee a local optimal, we examine the consequence of different iteration numbers. Table \ref{table:convmf} reports the quantitative result of using different iteration numbers, i.e. 1, 3, 5, 7, and 10. With increasing iterations, our proposed model performs better. However, we observe that no additional performance benefit can be gained when the iteration number exceeds 5. Fig. \ref{fig:ProbMap} presents the probability map during segmentation, where the light color represents the region with a lower probability while the dark color represents the area with a higher probability. We observe that using only one iteration, CANet can outline the region of interest using the fused feature maps. By increasing the iteration number to 3 or 5, CG-ACRF can gradually extract an optimal feature map, leading to accurate segmentation. We further increase the iteration number to 7 and 10 but no further improvement has been made. Therefore, we set the iteration number to 5 as a working trade-off between the segmentation performance and the number of the engaged parameters.

\subsection{Comparison with State-of-The-Art methods} We choose several State-of-The-Art deep learning based brain glioma segmentation methods, including 3D UNet \cite{cciccek20163d}, Attention UNet \cite{oktay2018attention}, PRUNet \cite{brugger2019partially}, NoNewNet \cite{isensee2018no} and 3D-ESPNet \cite{mehta20193despnet}. We first consider 5-fold cross-validation using the BraTS2017 training set. Each fold contains randomly chosen 228 cases for training and 57 cases for validation. In these cross-validation experiments, we consider CANet with CC+GC and CGA-CRF fusion modules with 5-iteration, leading to the best performance in the ablation tests. As shown in Table \ref{table:SOTACompare}, our CANet outperforms the other State-of-The-Art methods on several metrics while the results of the proposed method is competitive for the other metrics. The Dice score of CANet is 0.903 and 0.873 for the whole tumor and the tumor core respectively, where the former is 8\% higher and the later is 3\% higher than individual runner up results. The Hausdorff95 values of CANet are 3.569 and 4.036 for the whole tumor and the tumor core, which are much lower than the runner up scores, i.e. 4.156 and 5.778, respectively. Based on the individual score generated from the official evaluation server, we argue that the Dice score of ET from our proposed CANet is affected by the data-imbalance issue. The evaluation of enhancing tumor only considers the prediction of pertumoral edema, which only exists in the High-Grade Glioma (HGG) patients. As the training set contains more HGG cases than the LGG (Lower-Grade Glioma) cases, our system may learn a bias and make inaccurate prediction on some LGG cases in validation as the HGG cases contain false positives for the prediction of pertumoral edema labels. This false positive prediction leads to 0 Dice score on ET instead of 1 for LGG cases, thus decrease the performance of our system. We further summarise additional and detailed data-imbalance issues and failure case analysis in Supplementary G.

To further evaluate the segmentation output, we compare the segmentation output of the proposed approach against the ground-truth. Fig. \ref{fig:SOTA} shows that the proposed CANet can effectively predict the correct regions including small tumor cores and complicated edges while the other state of the art methods fail to do so. In Supplementary D, Fig. S4 presents the exemplar segmentation result and the ground-truth annotation in 3D visualisation. From Supplementary D Fig. S4, we observe that our proposed CANet effectively captures 3D forms and shape information in all different circumstances.

Fig. S5 of Supplementary E reports the training curve of CANet and the other State-of-The-Art methods. Our proposed method converges to a lower training loss using less epochs, compared against the other methods. Taking the advantage of the powerful feature interaction graph and the proposed fusion module CGA-CRF, CANet achieves satisfactory outlining of brain glioma. With the training epoch increasing, CANet fine-tunes the segmentation map and successfully detects small tumor cores and boundaries. We illustrate the probability map of our proposed CANet and 3D UNet in Fig. \ref{fig:probComp}. From Fig. \ref{fig:probComp}, we witness that our CANet can localize the shape contour of the target tumor to achieve precise segmentation, while the standard 3D UNet may lead to uncertainty, e.g. first row (WT probability map) and last row (TC probability map) in Fig. \ref{fig:probComp}. Also the standard U-Net may misclassify healthy surroundings to be tumor tissues, e.g. second row (WT probability map) and thrid row (ET probability map) in Fig. \ref{fig:probComp}.\\

We further investigate the segmentation results on the BraTS2017, BraTS2018 and BraTS2019 validation sets, where the quantitative result of each patient case is generated from the online evaluation server. The mean and standard deviation results are reported in Tables \ref{table:17val}, \ref{table:18val} and  \ref{table:19val}. Box plot in Supplementary F - Fig. S6 shows the distribution of the segmentation result among all the patient cases in the validation set. For the BraTS2017 validation set, our proposed CANet with complete CC+GC and 5-iteration CGA-CRF achieves the State-of-The-Art results of mean Dice scores on ET, TC and mean Hasdorff95 score on ET among the single model segmentation benchmarks. Our CANet has the Dice on ET of 0.728 with standard deviation 0.286, higher than the approach reported in \cite{pereira2019adaptive}. The Dice on TC by CANet is 0.821, which is higher than the runner-up result reported in \cite{roy2018recalibrating}. The Hausdorff95 on ET of CANet is 5.496, which is much lower than the runner-up generated in \cite{pereira2019adaptive}. For the BraTS2018 validation set, our proposed CANet achieves the State-of-The-Art result for Hausdorff95, i.e. 7.674, on the tumor core, while the other results are all runner-ups. Note that the method proposed by Myronenko \cite{myronenko20183d} has the best performance using most of the evaluation metrics. In the Myronenko's method, they set up an additional branch of using autoencoder to regularise the encoder backbone by reconstructing the input 3D MRI image. This autoencoder branch greatly enhances the feature extraction capability of the backbone encoder. In our framework, we regularise the network weights using a L2-regularisation without any additional branch, and the result of our proposed CANet is better than the other single prediction methods. Be reminded that the standard single prediction models generate the segmentation outcomes only using one network, and do not need much computational resources and a complicated voting scheme. For BraTS2019, our proposed CANet with complete CC+GC and 5-iteration CGA-CRF achieves competitive performance against the top performer. CANet's Dice on TC reaches the runner-up and Dice on ET reaches the third place, compared with the other  State-of-The-Arts methods \cite{jiang2019two, zhao2019bag, wang20193d, li2019multi, myronenko2019robust}. Note that methods like Jiang et al. \cite{jiang2019two} used two U-Net in the their single architecture where the first U-Net generates the coarse segmentation result and the second U-Net refines the coarse result to a precise one for the final segmentation output. Thus, this method can be regarded as a variant of model ensembling. Compared with the other State-of-The-Art methods, the result of our proposed CANet is very competitive in terms of accuracy.

Note that even adding additional neural blocks for mean field approximation, the parameters of our system suit the system well. We report the parameter size of our proposed model and other baseline candidates in Tables \ref{table:17val} and \ref{table:18val} (we employ the parameter setup of \cite{zhang2020exploring}). Our system maintains the parameter size at a middle level (3.34E7), compared with the other baseline methods such as Islam et al. \cite{islam2017multi} (1.38E8), Shaikh et al. \cite{shaikh2017brain} (2.31E7), Castillo et al. \cite{castillo2017volumetric} (2.22E7) and Carver et al. \cite{carver2018automatic}(2.20E7). We train our system for 200 epoches with a batch size of 2, which takes 27 hours for training. For the evaluation purpose, our system carries out each case within 0.88 seconds.

\section{Conclusion}
\label{Conclusion}
In summary, we have proposed a novel 3D MRI brain glioma segmentation approach called CANet. Considering different contextual information with standard and graph convolutions, we proposed a novel hybrid context aware feature extractor combined with deep supervised convolution and graph convolution contexts. Different from previous works that used naive feature fusion schemes such as element-wise summation or channel-wise concatenation, we here designed a novel feature fusion model based on conditional random fields, called context guided attentive conditional random field (CGA-CRF), which effectively learns the optimal latent features for downstream segmentation. Furthermore, we formulated the mean-field approximation within CGA-CRF as a convolutional operation, which incorporates the CGA-CRF in a segmentation network to perform end-to-end training. We conducted extensive experiments to evaluate the effectiveness of the proposed feature interaction graph method, CGA-CRF and the complete CANet framework. The results have shown that our proposed CANet achieved the State-of-The-Art results with several evaluation metrics. In the future, we consider combining the proposed network with novel training methods that can better handle the imbalance issue in the datasets.

\bibliographystyle{IEEEtran}
\bibliography{tmi}

\clearpage
\onecolumn

\setcounter{table}{0}
\setcounter{figure}{0}
\setcounter{equation}{0}
\setcounter{page}{1}
\renewcommand\thefigure{S\arabic{figure}}
\renewcommand\thetable{S\arabic{table}}


\section{Supplementary B}
\begin{figure*}[hbt]
    \centering
    \includegraphics[width=0.88\textwidth]{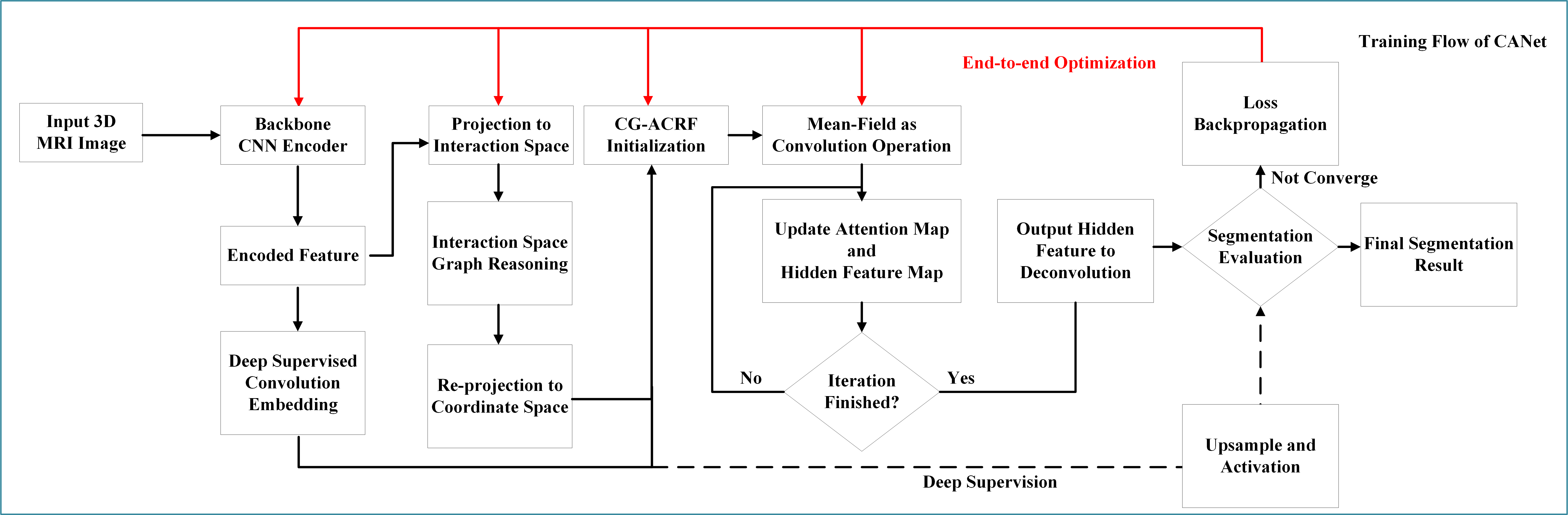}
    \caption{Training flow of the proposed CANet. Best viewed in colors.}
    \label{fig:trainingflow}
\end{figure*}

Fig. \ref{fig:trainingflow} summarises the training steps of CANet.

\section{Supplementary C}
\textbf{Data Augmentation} For each sequence in each case, we set all the voxels outside the brain to zero and normalise the intensity of the non-background voxels to be of zero mean and unit variance. During the training, we use randomly cropped images of size 128$\times$128$\times$128.
We further set up a common augmentation strategy for each sequence in each case: (i) Randomly rotate an image with the angle between [-20$^{\circ}$, +20$^{\circ}$]; (ii) Randomly scale an image with a factor of 1.1; (iii) Randomly mirror flip an image across the axial coronal and sagittal planes with the probability of 0.5; (iv) Random intensity shift between [-0.1, +0.1]; (v) Random elastic deformation with $\sigma = 10$.

\textbf{Implementation Details} We implement the proposed CANet and other benchmark experiments using the PyTorch framework and deploy all the experiments on 2 parallel Nvidia Tesla P100 GPUs for 200 epochs with a batch size of 4. We use the Adam optimizer with an initial learning rate $\alpha_0 = 1\mathrm{e}{-4}$. The learning rate is reduced by a factor of 5 after 100, 125 and 150 epochs. We use a $L$2 regulariser with a weight decay of $1\mathrm{e}{-5}$. We store the weights for each epoch and use the weights that lead to the best dice score for inference. The source code will be publicly accessible\footnote{https://github.com/ZhihuaLiuEd/canetbrats}.

\section{Supplementary D} 
\begin{figure*}[hbt]
    \centering
    \includegraphics[width=0.4\textwidth]{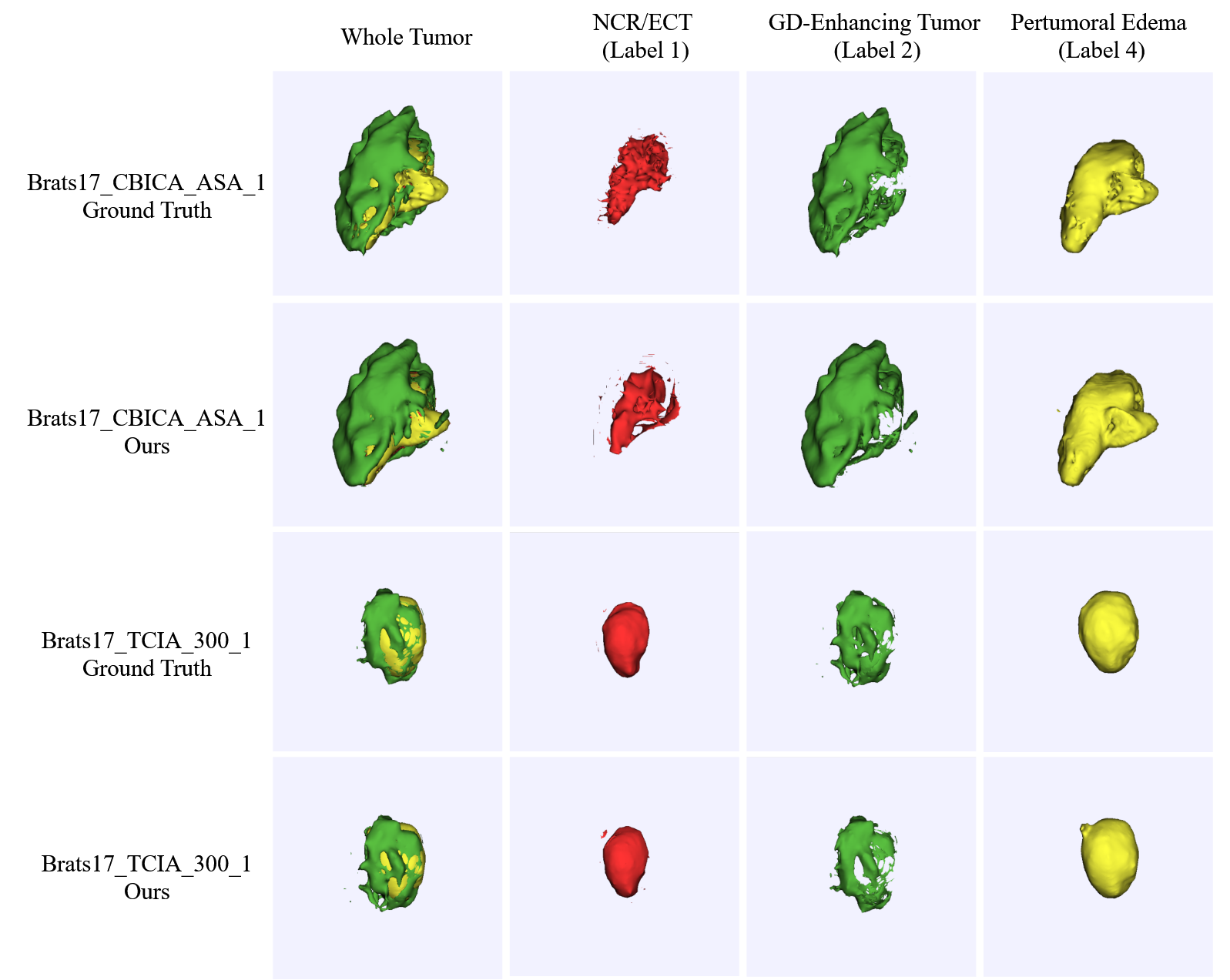}
    \caption{3D segmentation results of two volume cases by cross validation on the BraTS2017 training set. The first and third rows indicate the ground truth annotations. The second and fourth rows indicate the segmentation result of our proposed CANet with HCA-FE and 5-iteration CG-ACRF, respectively. Rows from left to right indicate the qualitative comparison of the whole tumor, NCR/ECT, GD-enhancing tumor and Pertumoral Edema respectively. Best viewed in colors.}
    \label{fig:3D}
\end{figure*}

Fig. \ref{fig:3D} shows the exemplar segmentation result and the ground truth annotation in 3D visualisation described in Section \Romannum{5}-B.

\section{Supplementary E} 
\begin{figure*}[hbt]
    \centering
    \includegraphics[width=0.5\textwidth]{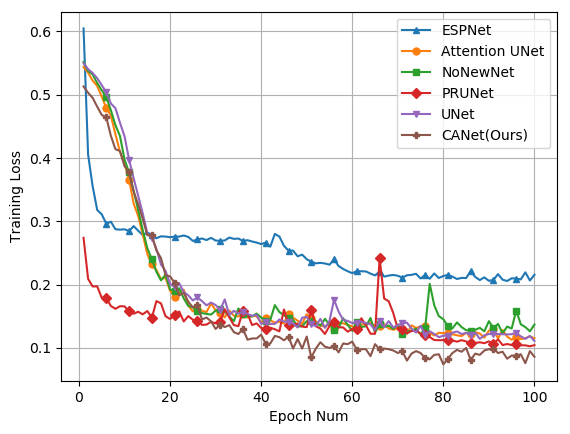}
    \caption{The learning curve of the state of the art methods and our proposed CANet with HCA-FE and 5-iteration CG-ACRF. Best viewed in color.}
    \label{fig:training record}
\end{figure*}

Fig. \ref{fig:training record} reports the training curve of CANet and the other state-of-the-art methods using the BraTS2017 training set, described in Section \Romannum{5}-B.

\section{Supplementary F} 
\begin{figure*}[hbt]
    \centering
    \includegraphics[width=0.45\textwidth]{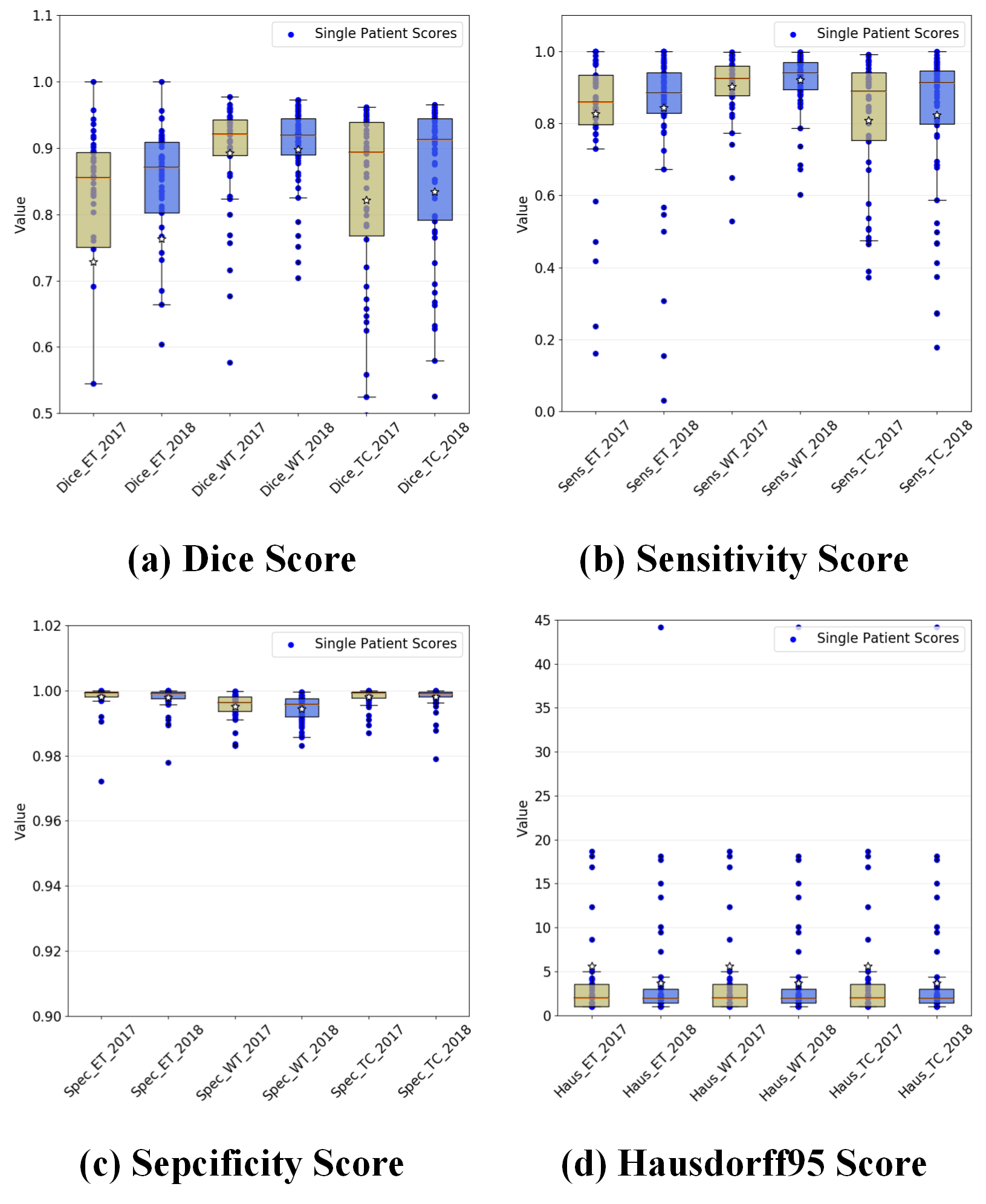}
    \caption{Boxplot of the segmentation results by CANet with HCA-FE and 5-iteration CG-ACRF, respectively. Dots within yellow boxes are individual segmentation results generated for the BraTS2017 validation set. Dots within blue boxes are individual segmentation results generated for the BraTS2018 validation set. Best viewed in color.}
    \label{fig:boxplot}
\end{figure*}

Fig. \ref{fig:boxplot} shows the distribution of the segmentation results among all the patient cases in the BraTS2017 and BraTS2018 validation sets described in Section \Romannum{5}-B.
\clearpage

\section{Supplementary G} 
\begin{figure*}[hbt]
    \centering
    \includegraphics[width=\textwidth]{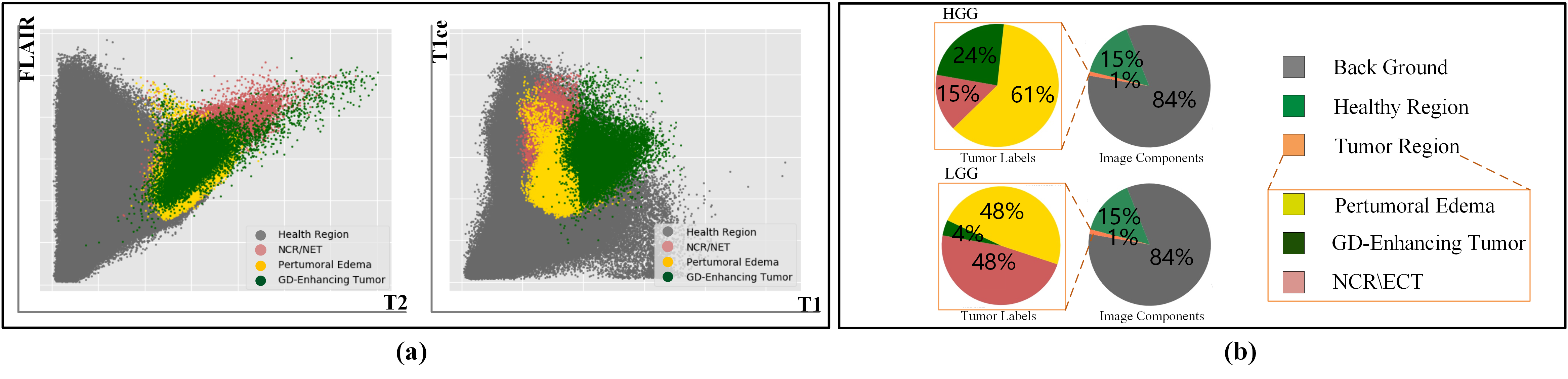}
    \caption{Statistics of the BraTS2017 training set. The left hand side figure of (a) shows the FLAIR and T2 intensity projection, and the right hand side figure shows the T1ce and T1 intensity projection. (b) is the pie chart of the training data with labels, where the top figure shows the HGG data labels while the bottom figure shows the LGG labels. There are large regions and label imbalance cases here. Best viewed in colors.}
    \label{fig:imbalance}
\end{figure*}

\begin{figure*}[hbt]
    \centering
    \includegraphics[width=\textwidth]{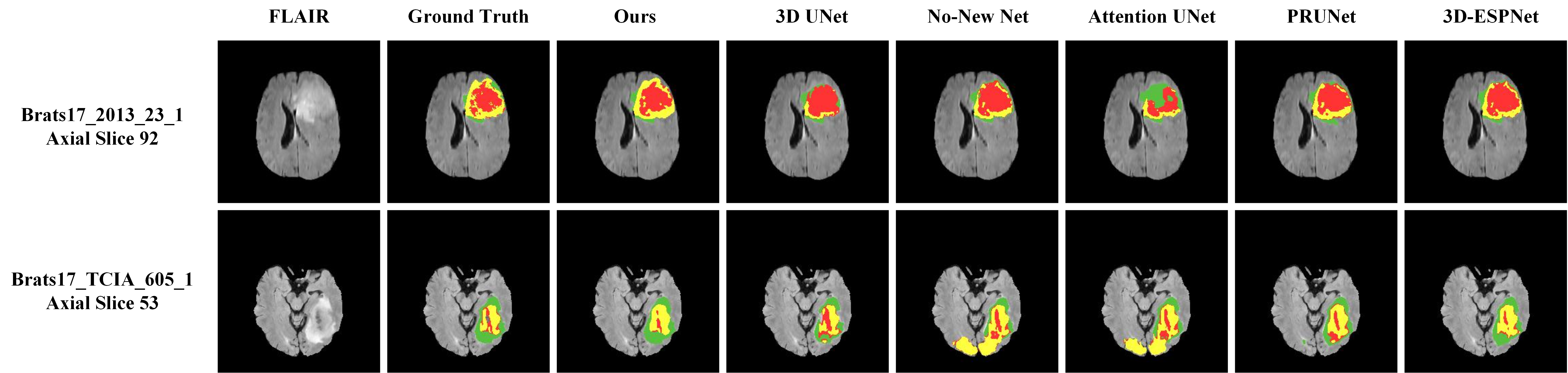}
    \caption{Qualitative comparisons in the failure cases. Rows from left to right indicate the input data of the FLAIR modality, ground truth annotation, segmentation result from our CANet, segmentation result from the other SOTA methods respectively. Our results look better than the SOTA methods' results. Best viewed in colors.}
    \label{fig:FailureCase}
\end{figure*}

Fig. \ref{fig:imbalance} shows the statistical information of the BraTS2017 training set. As an example, we here report two failure segmentation cases by our proposed approach, shown in Fig. \ref{fig:FailureCase}. During the whole training process, CANet focuses on extracting feature maps with different contextual information, e.g. convolutional and graph contexts. However, we have not designed specific strategies for handling the imbalanced issue of the training set. The imbalanced issue is presented in two aspects. Firstly, there exists an unbalanced number of voxels in different tumor regions. As the exemplar case named ``Brats17\_TCIA\_605\_1" is shown in Fig. \ref{fig:FailureCase}, the NCR/ECT region is much smaller than the other two regions, suggesting poor performance of segmenting NCR/ECT. Secondly, there exists an unbalanced number of patient cases from different institutions. This imbalance introduces an annotation bias where some annotations tend to connect all the small regions into a large region while the other annotation tends to label the voxels individually. As the exemplar case named "Brats17\_2013\_23\_1" is shown in Fig. \ref{fig:FailureCase}, the ground truth annotation tends to be sparse while the segmentation output tends to be connected together. In the future work, we will consider an effective training scheme based on active/transfer learning which can effectively handle the imbalance issue in the dataset. In spite of the imbalance issue, {\bf our segmentation method on the overall cases qualitatively outperforms the other state-of-the-art methods}.

\end{document}